\RequirePackage{tikz}
\documentclass{article}

\usepackage[T1]{fontenc}
\usepackage{amsmath}
\usepackage{physics}
\usepackage{amsmath}
\usepackage{tikz}
\usepackage{mathdots}
\usepackage{yhmath}
\usepackage{cancel}
\usepackage{color}
\usepackage{siunitx}
\usepackage{array}
\usepackage{multirow}
\usepackage{amssymb}
\usepackage{textcomp}
\usepackage{gensymb}
\usepackage{tabularx}
\usepackage{extarrows}
\usepackage{booktabs}
\usetikzlibrary{fadings}
\usetikzlibrary{patterns}
\usetikzlibrary{shadows.blur}
\usetikzlibrary{shapes}

\usepackage{mathdots}
\usepackage{yhmath}
\usepackage{cancel}
\usepackage{color}
\usepackage{siunitx}
\usepackage{array}
\usepackage{multirow}

\usepackage{natbib}
\usepackage{caption}

\usepackage{booktabs, multirow} 
\usepackage{soul}
\usepackage{xcolor} 
\usepackage{changepage,threeparttable}
\usepackage{lscape}
\usepackage{tabularx}
\usepackage{pgfplots}

\usepackage{longtable}

\pgfplotsset{compat=1.17}

\usetikzlibrary{positioning, shapes, shadows, arrows}
\usetikzlibrary{decorations.pathreplacing}
\tikzstyle{abstract}=[circle, draw=black, fill=white]
\tikzstyle{labelnode}=[circle, draw=white,opacity=.2,text opacity=1]
\tikzstyle{invisiblenode}=[circle,dashed, inner sep=1pt,circle split,line width=1mm,minimum size=1.5cm]
\tikzstyle{line} = [draw, -latex']

\newcounter{IEEE_VER}
\title{Toward cross-subject and cross-session generalization in EEG-based emotion recognition: Systematic review, taxonomy, and methods 
{\footnotetext{This work has been published on \textit{Neurocomputing} journal. Please refer to the final version of the paper on \url{https://doi.org/10.1016/j.neucom.2024.128354}. Old title "Machine Learning Strategies to Improve Generalization in EEG-based Emotion Assessment: a Systematic Review" has been changed to the current one.}}

}

\author{Andrea~Apicella$^{1,4}$, Pasquale Arpaia$^{1,4}$, Giovanni D'Errico$^2$,\\ Davide Marocco$^3$, Giovanna Mastrati$^{1,4}$, Nicola Moccaldi$^{1,4}$,\\ Roberto Prevete$^{1,4}$\\
        \small $^1$Laboratory of Augmented Reality for Health Monitoring (ARHeMLab),\\
        \small$^2$ Department of Applied Science and Technology, Polytechnic University of Turin,\\
        \small$^3$ Natural and Artificial Cognition Laboratory,\\
        \small$^4$ Department of Electrical Engineering and Information Technology, University of Naples Federico II \\
        }

\date{}

\usepackage{natbib}
\usepackage{graphicx}

\begin{document}

\maketitle
\begin{abstract}
A systematic review on machine-learning strategies for improving generalizability (cross-subjects and cross-sessions)
electroencephalography (EEG) based in emotion classification was realized. 
In this context, the non-stationarity of EEG signals is a  critical issue and can lead to the \textit{Dataset Shift} problem. Several architectures and methods have been proposed to address this issue, mainly based on transfer learning methods.
418 papers were retrieved from the \textit{Scopus}, \textit{IEEE Xplore} and \textit{PubMed} databases through a search query focusing on modern machine learning techniques for generalization in EEG-based emotion assessment. Among these papers, 75 were found eligible based on their relevance to the problem. Studies lacking a specific cross-subject and cross-session validation strategy and making use of other biosignals as support were excluded.
On the basis of the selected papers' analysis, a taxonomy of the studies employing Machine Learning (ML) methods was proposed, together with a brief discussion on the different ML approaches involved.
The studies with the best results in terms of average classification accuracy were identified, supporting that transfer learning methods seem to perform better than other approaches. A discussion is proposed on the impact of (i) the emotion theoretical models and (ii) psychological screening of the experimental sample on the classifier performances.
\end{abstract}
\section{Introduction}
Emotions are our internal compass and play a primary role in learning, reasoning, decision-making processes, and communication between individuals. The Information and Communication Technology (ICT) sector's interest in emotions has grown tremendously in recent years, shaping the concept of \textit{affective computing}, an emerging field aimed at monitoring and predicting emotions in order to improve human-computer interaction \cite{cambria2017affective}; for instance, the introduction of \textit{affective loops} makes it possible to implement increasingly adaptive human-machine interfaces and virtual assistants tailored to users \cite{saganowski2020emotion}, or the outputs of emotion monitoring systems, in the healthcare context, can be useful in the treatment of psychological disorders based on emotional deficits, in autism \cite{feng2018wavelet}, in the improvement of wellbeing  \cite{healy2018machine}, and in stress containment \cite{saganowski2022bringing}.

\begin{figure*}
\begin{center}
  \includegraphics[width=0.8 \textwidth]{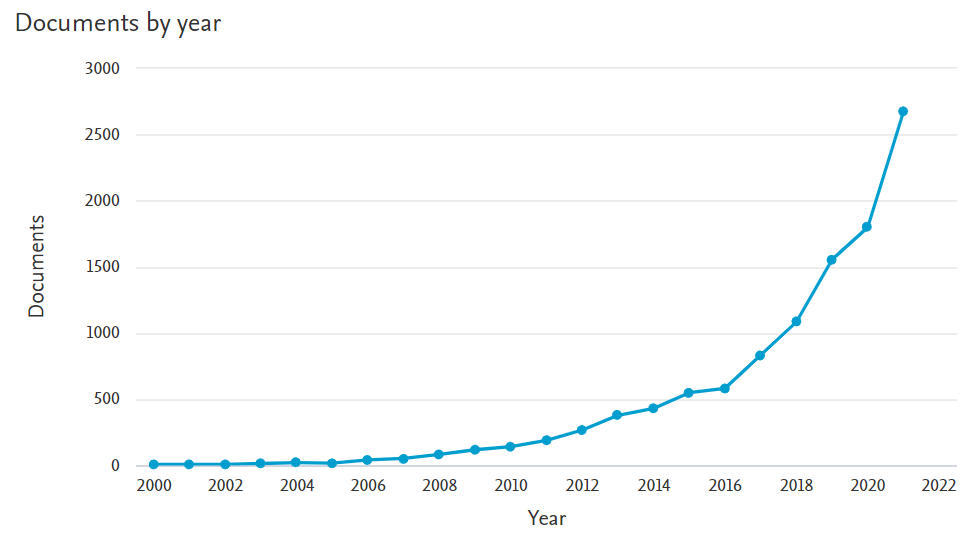}\\
  \caption{Scopus trend for EEG-based Emotion Recognition studies.}
  \label{trend}
\end{center}
\end{figure*}

In particular, in this context, there is a growing interest in the literature for Brain-Computer Interface (BCI) systems based on EEG signals \cite{torres2020eeg}. 
In fact, the number of annual scientific publications indexed on \textit{Scopus} database on the topic of EEG-based emotion recognition shows an exponential growth trend (see Fig. \ref{trend}).


A critical issue underlying the processing and classification of EEG signals is their inherent variability among different subjects or different acquisition times (i.e. sessions) of the same subject, since the EEG signal is usually stochastic and stationary only for short intervals (generally ranging from a few seconds to minutes) \cite{inouye1995new, im2018computational, sornmo2005bioelectrical}. More in detail, the EEG signal is not a Wide Sense Stationary signal \cite{cao2011application}. This characteristic of non-stationarity implies a variation in the temporal and spectral characteristics of the EEG signal over time. 
This is an open issue in the literature leading to a loss of generalizability for classification systems across subjects (\textit{inter-subject} task) and, for the same subject, across different sessions (\textit{intra-subject} task) \cite{nasiri2020attentive}. 

Data-driven approaches using Machine Learning (ML) are often employed at multiple levels in the EEG signal processing pipeline to pursue the classification of emotional states and their generalization across subjects and sessions.


Currently, the literature shows increasing use of modern machine learning strategies, adopting deep neural networks and transfer learning-based approaches, such as domain adaptation, domain generalization and/or hybrid methods \cite{zhao2021plug}. 
This paper proposes a systematic review on the use of machine learning to improve generalizability capabilities in EEG-based emotion recognition systems across different subjects and sessions.

As will be discussed in detail in the next section, 
several surveys have been proposed in recent years, gathering and discussing the main directions of the literature on this research topic.
However, to the best of our knowledge, a focus on the application of ML methods to improve the inter/intra-subjective generalization performance of EEG-based emotion recognition is missing in the literature. 

The rest of the paper is organized as follows: 
Section \ref{related_works} reviews related works, with reference to recent surveys carried out on this specific topic. Section \ref{background} presents a theoretical background on EEG, with a first part focused on BCIs for emotion recognition and a second part on ML for emotion recognition. Section \ref{method} presents the used search queries and the paper selection process according to the PRISMA method \cite{liberati2009prisma}. Section \ref{results} presents the results of the review, proposing a taxonomy of the ML methods currently proposed in the selected papers, discussing the ML methods with respect the proposed taxonomy. A statistical analyses of the results was reported. Section \ref{discussion} aims to discuss the results obtained, reporting the most promising lines of research and approaches that have emerged and highlighting possible future directions in this area. Finally, Section \ref{conclusion} draws conclusions.

\section{Related Works}
\label{related_works}

Several reviews have been conducted in recent years on this line of research. Alarcao and Fonseca \cite{alarcao2017emotions} focus on the generic topic of EEG-Emotion Recognition, presenting a review of papers published in the period from 2009 to 2016. The survey appears interesting in focusing on the different stages of the emotion recognition process from EEG signals and proposing a criterion for assessing the quality of the papers by applying a set of well-known guidelines (Brouwer's recommendations \cite{brouwer2015using}). However, there is no in-depth analysis on the issue of inter/intra-subject generalisation, nor the  EEG-nonstationarity problem is addressed. Other reviews \cite{lotte2018review, wu2020transfer} analyse works about the EEG-based classification methods, but without focusing on the emotion domain. Wu et al. \cite{wu2020transfer}, offer a non-systematic review focusing on the affective BCIs (aBCIs), but without an in-depth analysis on the emotion recognition problem. The study proposed in \cite{suhaimi2020eeg} deals with the EEG-inter/intra-subject variability problem as a specific topic on which to focus future research efforts, without exploring the problem in-depth. Recently, Li and colleagues \cite{li2021can} published a review focusing on the topic of EEG-based emotion recognition and discussing the importance of transfer learning. While offering some interesting results, it does not present itself as a systematic review (only 18 studies were reported without PRISMA methodology to collect them). 
Thus, differently from the cited works, a systematic literature review focused on the  inter/intra-subject generalization on EEG-based emotion recognition systems and the use of modern ML-based methods as a possible solution is proposed in this paper. 

\section{Theoretical Background}
\label{background}

\subsection{BCI for Emotion Recognition}
The emotional states can be recognized through several biosignals. In particular, brain signals have received increasing attention from the scientific community. In fact, the EEG signal resulted particularly effective for emotion recognition due to the high temporal resolution and its non-invasiveness. The EEG signal has a frequency range between [0.01, 100.00] $Hz$, an amplitude 
varying typically within the range [-100, 100] $\mu V$, and a power spectral density higher at lower frequencies \cite{daly2012does}.
Five background rhythms are present in the EEG and can be classified into different frequency bands: delta [0.5, 4.0] $Hz$, theta [4, 7] $Hz$, alpha [8, 13] $Hz$, beta
[14, 30] $Hz$, and gamma [30, 100] $Hz$. 

The 10–20 International Positioning System is an internationally recognized method to place the electrodes on the scalp \cite{acharya2016american} for the EEG signal recording.
The method allows to maintain a standardized EEG electrodes placement proportional to the scalp size and shape in order to preserve the relationship between each location and the underlying brain area.
In order to obtain a high-quality EEG, a substantial requirement is the use of high performance electrodes. The electrodes need to ensure a good and constant electrical contact with the skin and therefore need low impedance properties \cite{casson2019wearable}. The electrode-skin contact can either be ensured by adding a conductive gel between the electrode and the skin or by increasing the contact surface that ensures electrical contact.
Recently, besides wet electrodes, dry electrodes are employed for the EEG signal recording. A good signal quality and comparable performances with respect to wet electrodes are achieved using dry electrodes \cite{lopez2014dry}.

Besides the quality of the EEG signal, the emotion induction methods and the eliciting stimuli employed represent a crucial point for the effectiveness of the emotional elicitation.
Facial and body movements, recall of past events, odors, images, film clips, and music are techniques currently used in laboratories for inducing emotions.
Current literature reports that film clips, images, and music are particularly effective to elicit emotions \cite{westermann1996relative,lang2005international}. The use of images over other kind of stimuli represents a great advantage insofar as the images are standardized stimuli. Datasets of images were experimentally validated (e.g., International Affective Picture System - IAPS \cite{lang2005international}, Open Affective Standardized Image Set - OASIS \cite{kurdi2017introducing}, and Geneva Affective Picture Database - GAPED \cite{dan2011geneva}). 
There are several publicly-available databases for EEG-based emotion recognition (e.g., DEAP \cite{koelstra2011deap}, SEED \cite{zheng2015investigating}, and DREAMER \cite{katsigiannis2017dreamer}). 
Each dataset is characterized by different physiological signals and a well-established experimental setup (in terms of stimulus sources, emotional theory adopted, number of subjects, and psychometric metrological references). For a comprehensive description of the various available datasets see \cite{li2021can}.

In case of self-produced EEG data, the preprocessing stage is fundamental to filter out the noise from the brain activity signal.
Some steps are often helpful to achieve a successful EEG signal preprocessing: (i) line noise removal, (ii) referencing, (iii) bad channels removal, and (iv) artifacts removal. They can be summarized as:
\begin{itemize}
    \item Line noise removal: line noise is an artifact which contaminates the gamma band of the EEG signal, specifically at 50 $Hz$ (60 $Hz$ in the USA) \cite{im2018computational}. Conventional filters such as lowpass filters with cutoff frequencies between 50 $Hz$ and 70 $Hz$ can be used to remove this artifact and reduce the noise at higher frequencies, notch filters must be employed to reject the 50 $Hz$ power supply.
\item Referencing: when acquiring the EEG signals, the voltage in a specific scalp area is measured with respect to a reference electrode.
If the reference electrode is subjected to artifacts, all electrodes are also affected.
The re-referencing allows to minimize the impact of the reference electrode by subtracting a reference channel from the original EEG channels.
Most used reference electrodes are the mastoid channel, the EEG signal at a specific channel, the average of the mastoid channels, or the average of all EEG channels \cite{lepage2014statistically}.
To avoid lateralization effects \cite{lepage2014statistically}, also Cz and FCz electrodes are often used as references.
\item Bad channels removal: EEG signal of one or more channels can be also contaminated by the noise due to a poor contact between the electrode and the scalp.
Therefore, it is necessary to detect the noisy or bad channels and remove them.
Methods to find out bad channels are the visual inspection of each single channel, the dispersion and the correlation criteria \cite{da2018automatic}.
\item Artifacts removal: artifacts are all the unwanted signals that may affect the measurement and corrupt the EEG signal. They can be due to all the physiological systems different from the brain, such as heart, eyes, muscle, etc., or all the environmental noise, such as wireless signals, electrode adhesion, cable movements, etc.
In the EEG signals, artifacts are present in specific frequency bands: ocular artifacts and cardiac activity are dominant below 4 $Hz$, muscle movements above 30 $Hz$.
Other physiological artifacts are due to skin perspiration, sweating, movements of the tongue, chest movements, etc. 
Removing artifacts from the EEG signal means to correct (or eliminate) the fluctuations introduced by the artifacts without causing distortions in the brain signal.
Part of the artifacts can be removed with the filtering process (e.g., line noise) but, for a more effective removal of artifacts, further processing is required \cite{jiang2019removal}.
Main artifacts removal methods are: linear regression, filtering, wavelet transform, empirical mode decomposition (EMD), Independent Component Analysis (ICA), Principal Component Analysis (PCA), Canonical Correlation Analysis (CCA), and Artefact Subspace Reconstruction (ASR). 
\end{itemize}

Once the EEG signal has been preprocessed, it is usually divided into epochs, and a feature extraction process is then applied.
EEG features can be categorized into three domains, namely time, frequency and time-frequency.\\
\begin{itemize}
    \item Time domain: the main features are the statistics of the signal, such as mean, variance, skewness, kurtosis, etc \cite{geethanjali2012time,vidaurre2009time,yuen2009classification}. Other time-domain features are the Hjorth parameters, namely Activity, Mobility, and Complexity \cite{oh2014novel}.
Good results in the recognition of emotional states can be achieved by using entropy-based features, i.e., approximate, sample, differential, and wavelet entropy \cite{phadikar2019survey}.
Higher-order crossing (HOC), the fractal dimension, and the Non-Stationary Index (NSI) \cite{patil2016feature,nan1988fractal,kroupi2011eeg} are further time domain feature often used for the EEG analysis.
\item Frequency domain: the most used feature is the power spectral density (PSD).
PSD is the signal power in the unit frequency band \cite{wang2011eeg}.
Other representative features of different emotional states involving the PSD are: (i) logarithm, (ii) maximum, (iii) minimum and, (iv) standard deviation of the power spectrum. 
\item Time-frequency domain: the time-frequency analysis (TFA) allows to observe spectrum changes with time\cite{zhang2019spectral}.
The short-time Fourier transform (STFT),the continuous wavelet transform (CWT), the discrete wavelet transform (DWT) \cite{hernandez2018detecting}, matching pursuit and empirical mode decomposition are the most used methods to extract time-frequency features. 
\end{itemize}

Often, the number of EEG features is very high, therefore a feature selection strategy is required \cite{jenke2014feature}.
Another critical point is represented by the large number of EEG channels often used for the signal acquisitions. A high number of channels can lead to high computational complexity. Therefore, the selection of the most informative EEG channels can be crucial \cite{9940267}.



\subsection{Machine Learning for Emotion Recognition}

\tikzset{every picture/.style={line width=0.75pt}} 
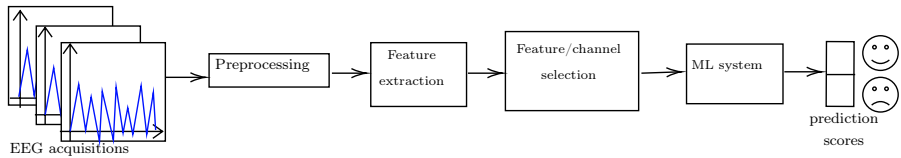
\begin{figure}
\scalebox{0.7}{
\begin{tikzpicture}[x=0.75pt,y=0.75pt,yscale=-1,xscale=1]

\draw  [color={rgb, 255:red, 0; green, 0; blue, 0 }  ,draw opacity=1 ][fill={rgb, 255:red, 255; green, 255; blue, 255 }  ,fill opacity=1 ] (4.5,21) -- (79.5,21) -- (79.5,92) -- (4.5,92) -- cycle ;
\draw    (91.5,72) -- (144.5,72) ;
\draw [shift={(146.5,72)}, rotate = 180] [color={rgb, 255:red, 0; green, 0; blue, 0 }  ][line width=0.75]    (10.93,-3.29) .. controls (6.95,-1.4) and (3.31,-0.3) .. (0,0) .. controls (3.31,0.3) and (6.95,1.4) .. (10.93,3.29)   ;
\draw    (563.5,68) -- (588.5,68) ;
\draw [shift={(590.5,68)}, rotate = 180] [color={rgb, 255:red, 0; green, 0; blue, 0 }  ][line width=0.75]    (10.93,-3.29) .. controls (6.95,-1.4) and (3.31,-0.3) .. (0,0) .. controls (3.31,0.3) and (6.95,1.4) .. (10.93,3.29)   ;
\draw    (460.5,69) -- (487.5,68.07) ;
\draw [shift={(489.5,68)}, rotate = 178.03] [color={rgb, 255:red, 0; green, 0; blue, 0 }  ][line width=0.75]    (10.93,-3.29) .. controls (6.95,-1.4) and (3.31,-0.3) .. (0,0) .. controls (3.31,0.3) and (6.95,1.4) .. (10.93,3.29)   ;
\draw   (265.5,47) -- (334.5,47) -- (334.5,93) -- (265.5,93) -- cycle ;

\draw   (493,48) -- (562.5,48) -- (562.5,91) -- (493,91) -- cycle ;
\draw   (149,55) -- (235.5,55) -- (235.5,79) -- (149,79) -- cycle ;
\draw    (237.5,69) -- (260.5,69) ;
\draw [shift={(262.5,69)}, rotate = 180] [color={rgb, 255:red, 0; green, 0; blue, 0 }  ][line width=0.75]    (10.93,-3.29) .. controls (6.95,-1.4) and (3.31,-0.3) .. (0,0) .. controls (3.31,0.3) and (6.95,1.4) .. (10.93,3.29)   ;
\draw  (5.5,87) -- (79.5,87)(12.9,24) -- (12.9,94) (72.5,82) -- (79.5,87) -- (72.5,92) (7.9,31) -- (12.9,24) -- (17.9,31)  ;
\draw [color={rgb, 255:red, 0; green, 20; blue, 255 }  ,draw opacity=1 ]   (11.9,86) -- (18.05,52.1) -- (23.05,85.1) -- (27.05,61.1) -- (33.05,93.1) -- (35.05,57.1) -- (43.05,92.1) -- (45.05,54.1) -- (50.05,87.1) -- (53.39,68.98) -- (57.05,88.1) -- (63.05,53.1) -- (67.05,88.1) -- (71.86,57.9) -- (73.5,85) ;
\draw  [color={rgb, 255:red, 0; green, 0; blue, 0 }  ,draw opacity=1 ][fill={rgb, 255:red, 255; green, 255; blue, 255 }  ,fill opacity=1 ] (24.5,35) -- (99.5,35) -- (99.5,106) -- (24.5,106) -- cycle ;
\draw  (23.5,99) -- (97.5,99)(30.9,36) -- (30.9,106) (90.5,94) -- (97.5,99) -- (90.5,104) (25.9,43) -- (30.9,36) -- (35.9,43)  ;
\draw [color={rgb, 255:red, 0; green, 20; blue, 255 }  ,draw opacity=1 ]   (30.9,99) -- (37.05,65.1) -- (42.05,98.1) -- (46.05,74.1) -- (52.05,106.1) -- (54.05,70.1) -- (62.05,105.1) -- (64.05,67.1) -- (69.05,100.1) -- (72.39,81.98) -- (76.05,101.1) -- (82.05,66.1) -- (86.05,101.1) -- (90.86,70.9) -- (92.5,98) ;
\draw  [color={rgb, 255:red, 0; green, 0; blue, 0 }  ,draw opacity=1 ][fill={rgb, 255:red, 255; green, 255; blue, 255 }  ,fill opacity=1 ] (42.5,47) -- (117.5,47) -- (117.5,118) -- (42.5,118) -- cycle ;
\draw  (41.5,111) -- (115.5,111)(48.9,48) -- (48.9,118) (108.5,106) -- (115.5,111) -- (108.5,116) (43.9,55) -- (48.9,48) -- (53.9,55)  ;
\draw [color={rgb, 255:red, 0; green, 20; blue, 255 }  ,draw opacity=1 ]   (48.9,111) -- (55.05,77.1) -- (60.05,110.1) -- (64.05,86.1) -- (70.05,118.1) -- (72.05,82.1) -- (80.05,117.1) -- (82.05,79.1) -- (87.05,112.1) -- (90.39,93.98) -- (94.05,113.1) -- (100.05,78.1) -- (104.05,113.1) -- (108.86,82.9) -- (110.5,110) ;
\draw   (613.73,46.3) -- (613.5,93) -- (593.6,92.9) -- (593.82,46.2) -- cycle ;
\draw    (593.5,70) -- (614.5,70) ;
\draw   (620,54.5) .. controls (620,47.6) and (625.71,42) .. (632.75,42) .. controls (639.79,42) and (645.5,47.6) .. (645.5,54.5) .. controls (645.5,61.4) and (639.79,67) .. (632.75,67) .. controls (625.71,67) and (620,61.4) .. (620,54.5) -- cycle ; \draw   (627.14,50.25) .. controls (627.14,49.56) and (627.71,49) .. (628.42,49) .. controls (629.12,49) and (629.69,49.56) .. (629.69,50.25) .. controls (629.69,50.94) and (629.12,51.5) .. (628.42,51.5) .. controls (627.71,51.5) and (627.14,50.94) .. (627.14,50.25) -- cycle ; \draw   (635.81,50.25) .. controls (635.81,49.56) and (636.38,49) .. (637.09,49) .. controls (637.79,49) and (638.36,49.56) .. (638.36,50.25) .. controls (638.36,50.94) and (637.79,51.5) .. (637.09,51.5) .. controls (636.38,51.5) and (635.81,50.94) .. (635.81,50.25) -- cycle ; \draw   (626.38,59.5) .. controls (630.63,62.83) and (634.88,62.83) .. (639.13,59.5) ;
\draw   (620,84.5) .. controls (620,77.6) and (625.71,72) .. (632.75,72) .. controls (639.79,72) and (645.5,77.6) .. (645.5,84.5) .. controls (645.5,91.4) and (639.79,97) .. (632.75,97) .. controls (625.71,97) and (620,91.4) .. (620,84.5) -- cycle ; \draw   (627.14,80.25) .. controls (627.14,79.56) and (627.71,79) .. (628.42,79) .. controls (629.12,79) and (629.69,79.56) .. (629.69,80.25) .. controls (629.69,80.94) and (629.12,81.5) .. (628.42,81.5) .. controls (627.71,81.5) and (627.14,80.94) .. (627.14,80.25) -- cycle ; \draw   (635.81,80.25) .. controls (635.81,79.56) and (636.38,79) .. (637.09,79) .. controls (637.79,79) and (638.36,79.56) .. (638.36,80.25) .. controls (638.36,80.94) and (637.79,81.5) .. (637.09,81.5) .. controls (636.38,81.5) and (635.81,80.94) .. (635.81,80.25) -- cycle ; \draw   (626.38,92) .. controls (630.63,88.67) and (634.88,88.67) .. (639.13,92) ;
\draw   (362.5,39) -- (458.5,39) -- (458.5,96) -- (362.5,96) -- cycle ;
\draw    (335.5,69) -- (358.5,69) ;
\draw [shift={(360.5,69)}, rotate = 180] [color={rgb, 255:red, 0; green, 0; blue, 0 }  ][line width=0.75]    (10.93,-3.29) .. controls (6.95,-1.4) and (3.31,-0.3) .. (0,0) .. controls (3.31,0.3) and (6.95,1.4) .. (10.93,3.29)   ;

\draw (4,118) node [anchor=north west][inner sep=0.75pt]   [align=left] {{\footnotesize EEG acquisitions}};
\draw (152,57) node [anchor=north west][inner sep=0.75pt]   [align=left] {{\footnotesize Preprocessing}};
\draw (578,96) node [anchor=north west][inner sep=0.75pt]   [align=left] {\begin{minipage}[lt]{40.35pt}\setlength\topsep{0pt}
\begin{center}
{\footnotesize prediction }\\{\footnotesize scores}
\end{center}

\end{minipage}};
\draw (267,50.95) node [anchor=north west][inner sep=0.75pt]  [font=\large] [align=left] {\begin{minipage}[lt]{39.89pt}\setlength\topsep{0pt}
\begin{center}
{\scriptsize Feature }\\{\scriptsize extraction}
\end{center}

\end{minipage}};
\draw (362,45.95) node [anchor=north west][inner sep=0.75pt]  [font=\large] [align=left] {\begin{minipage}[lt]{67.08pt}\setlength\topsep{0pt}
\begin{center}
{\scriptsize Feature/channel }\\{\scriptsize selection}\\
\end{center}

\end{minipage}};
\draw (495.01,57.85) node [anchor=north west][inner sep=0.75pt]   [align=left] {{\scriptsize ML system}};

\end{tikzpicture}}
\caption{A pipeline of a classical ML process involving EEG signals.}
\label{fig:pipeline}
\end{figure}
After that the EEG have been properly preprocessed and a proper set of features has been extracted, the data are ready to be fed to a supervised ML system. The typical pipeline of a ML framework applied to an EEG emotion recognition task is reported in Fig. \ref{fig:pipeline}.

In the current literature, a large part of research works proposed methods framed into Transfer Learning approach to tackle emotion recognition tasks. The motivation of this trend can be summarized as follows: in classical supervised ML a set of already labeled data has to be available. This implies that in the EEG emotion recognition tasks a set of EEG signals recorded from one or more subjects has to be mapped with the emotion felt during the acquisition. Labelled data can be then used to train the ML system, generating a ML model able to classify the input data. Once the ML model has been obtained, new unlabelled data can be fed to the ML model to obtain the predicted emotion/class. To evaluate the trained model, it is a good practice to reserve a part of the labelled data outside from the training stage. These data can then be used to evaluate the final model predictions using suitable performance metrics (e.g. accuracy). However, a standard hypothesis of the traditional ML methods states that all the available data, no matter if involved in the training process or not, come from the same probability distribution. Due to the characteristics of the EEG data, this assumption results not always verified in the EEG signal. Indeed, the EEG signals acquired from a subject can be strongly different from the one acquired from another subject, even under the same conditions \cite{im2018computational}. This can also happen for EEG signals acquired from the same subject in different times/sessions, leading to loss generalisation performances in cross-subject/session problems. In the current literature, this problem was initially addressed considering the availability of further unlabelled data belonging to the target subject/session which can help the training of the ML model (transductive learning approaches). However, these methods do not make any consideration about the data distributions. In fact, the training EEG data can be sensibly different in terms of probability distribution(s) from the data used outside the training stage. In ML literature, this can be viewed an instance of the Dataset Shift problem \cite{quinonero2008dataset}. In a nutshell, Dataset Shift arises when the standard ML assumption is not verified, so the distribution of the training data differs from the data distribution used outside of the training stage. The idea that the training data come from different probability distribution(s) respect to the data used outside of the training stage is the main hypothesis of the transfer learning approaches. 

In the last years, several architectures and methods have been proposed to address the dataset shift problem following the base assumptions of transfer learning, and different categorizations of the proposed methods have been reported \cite{pan2009survey, ganin2015unsupervised}. One of the first and most important review on Transfer Learning methods was proposed in \cite{pan2009survey}, however several new strategies (e.g., Domain Generalization-based works) have been proposed in the following years.

\section{Papers selection method}
\label{method}
\begin{figure*}
\begin{center}
  \includegraphics[width=0.8 \textwidth]{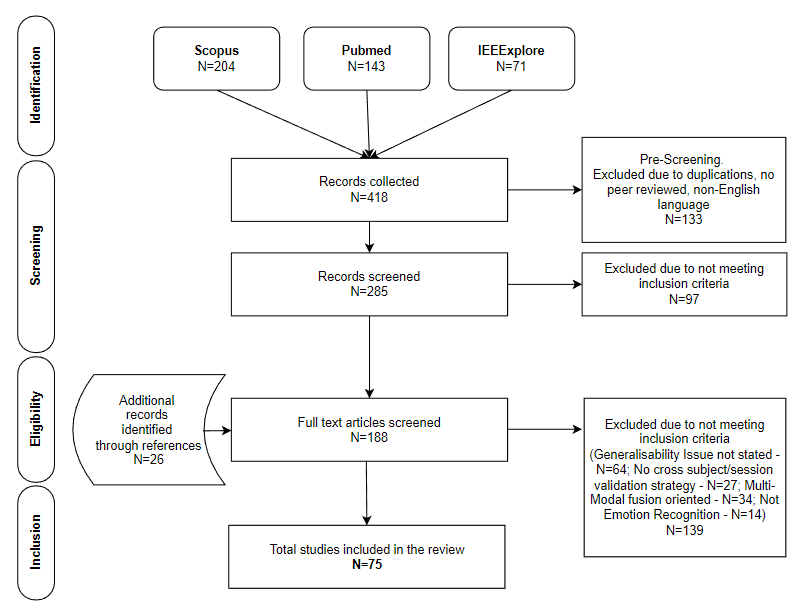}\\
  \caption{PRISMA flow diagram of the systematic
review process.}
  \label{PRISMA}
\end{center}
\end{figure*}

The present literature review took into account the guidelines for systematic literature reviews presented by Kitchenham (\cite{kitchenham2004procedures}), covering also the use of PRISMA (Preferred Reporting Items
for Systematic Reviews and Meta-Analyses) recommendations in order to transparently report the document extraction process (\cite{liberati2009prisma}). The survey was conducted covering the period between January 2010 to March 2022, using the following databases: \textit{Scopus}, \textit{IEEE (Institute of Electrical and Electronics Engineers)}, \textit{Xplore}, and \textit{PubMed}.

In accordance with PRISMA's recommendations, the pipeline consists in four successive steps: 'Identification', 'Screening', 'Eligibility' and, finally, 'Inclusion', which considerably narrowed down the amount of surveyed work. For the initial identification of the articles, the following query was used in all predicted data sources, taking into account titles and abstracts: EEG AND (Emotion OR Preference) AND ("Domain Adaptation" OR "Domain Generalisation" OR "Transfer Learning" OR "Adversarial" OR "Transfer" OR "Cross Session" OR "Cross Subject" OR "Cross Gender" OR "Non-stationary EEG").

From the first phase, 418 articles were collected. Then, as a first pre-screening process, the following criteria are used to exclude papers from the review: duplicated articles, not peer-reviewed articles, and not written in English articles.
For all articles that survived the screening stage, a careful examination of the full text was carried out. In a conclusive screening, the following papers were excluded: (a) all articles in which the problem of generalisability was not explicitly stated as a peculiar topic, (b) studies in which a cross-subject/session validation strategy was not clearly envisaged, (c) studies oriented towards a 'multimodal fusion', i.e. aimed at corroborating the EEG signal-based classification with other biosignals, (d) studies not specifically focusing on emotion recognition. Thus, 75 papers survived and were included in the review analysis. The complete flow diagram of the systematic review process using the PRISMA approach is presented in Fig. \ref{PRISMA}.

\section{Results}
\label{results}

\begin{figure*}
\begin{center}
  \includegraphics[width=0.8 \textwidth]{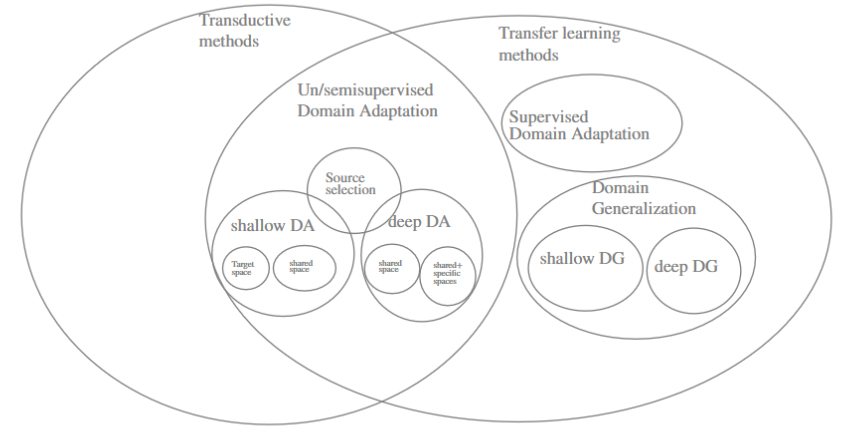}
  \caption{Proposed Taxonomy.}
  \label{taxo}
\end{center}
\end{figure*}

In this paper, a categorisation of the reviewed articles with respect ML methods which do not satisfy the standard ML hypothesis is proposed (see Figure \ref{taxo}). Among all the reviewed papers, a subset of them leveraging on the classical transductive learning approaches, i.e. considering a set of unlabelled data coming from the target subject/session in the training stage, is isolated. On the other side, the papers dealing with the cross-subject/session problem as an instance of the dataset shift problem were considered. These methods are known in literature as transfer learning methods, which can be further divided in: 
\begin{itemize}
\item \textbf{Unsupervised and Semisupervised Domain Adaptation} methods: they consider both the data coming from different subjects/sessions and unlabelled data coming from a target subject/session during the training stage. Therefore, they can be considered as an intersection between the transductive and transfer Learning methods;
\item \textbf{Supervised Domain Adaptation} methods: these methods rely on the hypothesis that both labelled and unlabelled data from the target subject/session are available during the training stage;
\item \textbf{Domain Generalization} methods: in these methods data from several probability distributions are available and can be used during the training, but no data from the target subject/session is used during the training stage.
\end{itemize}

On the basis of the above categorization shown in Fig. \ref{taxo}, in Tab. \ref{complete_table} all papers included in the review were reported, indicating if belonging to the proposed taxonomy or to classical ML methods. 
Moreover, for each research study several information is presented, such as the type of generalisation (cross-subject, cross-session, or cross-device), the EEG dataset, the adopted classifier (whether proposed as a personal contribution or adopted from the literature), and the validation strategy. Studies in which the description of the experimental setup was not sufficiently clear, especially in terms of validation strategies, were voluntarily omitted from the table for reproducibility issues.

In the following of this section, the reviewed papers are discussed in according with the Tab. \ref{complete_table}.

\subsection{\textbf{Classical machine learning approaches}}
A model trained on a set of EEG data acquired from a given subject at a specific time  (or during a specific session) could not work as expected in classifying EEG signal acquired  from  a different subject or from the same subject at different times. In other words, the model has poor generalization performance. To deal with this problem, several solution based on Machine Learning approach have been proposed over the years. The most common approaches proposed proper strategy of feature transformations or feature selection. The former wants to be transformations of the data able to hold only the most useful information with the hope that these are shared between all the subjects, while the latter are methods to select only useful information from the input signals without changing it. The feature extraction/selection is usually one of the first step of a machine learning pipeline, where the training data are transformed before being fed to the training stage of a machine learning model. These methods adopted the classical machine learning framework, that is that only the training data are available during the training stage, without any knowledge of the effective data on which the model will be effectively used. This can be viewed as a consequence of the the starting hypothesis of the traditional ML methods stating that all the available data, no matter if used in the training process or not, come from the same probability distribution. Under this hypothesis, the training data are enough to generalise over all the possible data. 

The underlying hypothesis is that a proper EEG data transformation is enough to allow a ML model to generalise well, independently from the fact the data belongs to the same subject/session used during the training or not. Going deeper, in a ML problem on EEG data the feature extraction and selection process can be made considering two different aspects:
i) the EEG signal or ii) the electrodes. In the first case, a proper transformation or selection strategy for the EEG signal is made. \\
The reviewed literature proposed different works that analyse if several known feature extraction methods are suitable to generalise across several datasets \cite{rayatdoost2018cross, li2018exploring}. In particular, in \cite{rayatdoost2018cross} the authors investigated the robustness of emotion recognition methods across different experimental conditions, subjects and datasets.

In \cite{yang2019multi,jiang2019cross} Sequential Backward Selection (SBS) was applied to find a good set of features that generalise across different subjects.
To find the best subset of features, SBS decreases the number of features in an iterative way measuring, at each step, the obtained accuracy on a given classifier (SVM in \cite{yang2019multi}, Decision Trees in \cite{jiang2019cross}). SBS method is adopted to exploit the significant differences between the classes. A leave-one-subject-out verification strategy was employed on DEAP and SEED datasets in \cite{yang2019multi}, while \cite{jiang2019cross} validates its results on DEAP and self-produced data.

In \cite{cai2019multiple,yin2017cross} a family of Transferable Recursive Feature Elimination (TRFE) methods are used to make a set of EEG features steadily distributed among all the training subjects, therefore removing the EEG features resulting not generic for all users. The proposed feature selector is validated using SVMs as classifiers on DEAP dataset both in within-subject and cross-subject ways.
In \cite{zhang2020eeg} Cross-subject Recursive Feature Elimination (C-RFE) is exploited to rank the features in  order of importance with the aim of removing the features giving a low  contribution to the classification. The method is validated on EEG data fed to SVMs.

In \cite{liu2020subject} an evolution of the well-known Differential Entropy features is proposed. The Dynamic Differential Entropy (DDE) features take into account also the time-domain instead of only the frequency domain extracted by the classical DE. The goal is to maximise the difference between classes minimising at the same time the difference within classes, learning a set of common characteristics across different subjects.

In \cite{li2019variational} a latent representation of the EEG data from SEED and DEAP is learned through a Variational Auto Encoder (VAE) and then classified using a LSTM. VAEs start from the hypothesis that all the data are generated by a random process involving latent variables. A VAE is usually trained to encode the input data into a latent representation and then mapping it to a reconstructed version of the data.  \cite{li2019variational} hypothesises that there exists learnable intrinsic features shared across several subjects EEG signals taking part in emotional processes.  These intrinsic features can be encoded by the VAE latent representations. The power of VAE to represent latent EEG factors is also investigated in \cite{li2020latent}, together with classical Auto-Encoders (AEs) and Restricted Boltzmann Machines (RBMs). Final classification is made with an LSTM and the generalisation performances are evaluated in LOSO mode on DEAP and SEED dataset.

In \cite{pandey2019subject} the cross-subject problem is tackled using Variational Mode Decomposition (VMD) as feature extraction technique. The system is validated in an hold-out way without any intersections between subjects' data in the training and the test set using a DNN as emotions classifier. Despite the encouraging results reported, no reason about why the proposed system works well in a cross-subject approach seems to be provided. 
In \cite{chen2021personal,fernandez2021cross,arevalillo2019combining} is shown that the normalisation scheme used to preprocess the EEG data can affect the cross-subject performances.

In \cite{chen2021personal} several normalisation methods were applied following two different schemes: i) All-subjects, where the whole dataset was normalised, ii) Single-subject, where the normalisation is made individually for each subject. The All-subject schema is the most common method used to mitigate the impact of each data values on the entire dataset. Single-subject, instead, consider each subject individually, applying normalisation on each subject. The authors empirically shown on SEED dataset that Single-subject Z-score performs better in a EEG emotion recognition problem respect to other normalisation schemes as min-max normalisation. On the same data, in \cite{fernandez2021cross} the authors apply single-subject $Z-score$ normalisation after each neural network layer (Stratified Normalisation).

In \cite{arevalillo2019combining} a simple transformation of the original data is proposed. It consists in using binary features having $0$ and $1$ as components values based on the fact that the feature is lower or higher than the median feature value. This leads to a more effective reduction of the subject-dependent part of the EEG signal.\\

Instead, considering the electrodes in place of the EEG signals, different channels selection strategies searching for a robust set of channels across the subjects was proposed in \cite{gupta2018cross,zhang2016relieff}.
To achieve EEG-based cross-session emotion recognition, in \cite{peng2021self} the author propose a way to learn the importance of the EEG channels and features to separate discriminative features from the noisy and redundant ones. The proposed strategy is evaluated on pairs of a-priori chosen sessions.

In \cite{tian2021personality} a neural network to classify emotion by EEG signals is proposed. The proposed model introduces a channel-attention layer to select the most important channels for a set of emotions. Notably, the different personalities across the subjects are taken into account, grouping together the subjects having similar personalities. Indeed, a different model for each group was trained. Validation is made on the ASCERTAIN dataset. This dataset results particularly suited for this task, since it links personality and emotional state with physiological reactions.

In other works, the structure of the electrodes is taken into account and modelled as a graph. 
Graph representation methodology resulted effective to model structured data achieving significant performance in many applications, included EEG emotion signal processing \cite{song2018eeg}. GNNs are useful to retain the spatial structure of the electrodes disposition.
Usually, the graph structure is fixed and given a priori following the spatial disposition of the electrodes on the scalp. Instead, Dynamical Graph Convolutional Neural Networks (DGCNN, \cite{song2018eeg}) and Self-Organized Graph Neural Network (SOGNN,  \cite{li2021cross}) organises the graph structure leveraging on the input brain signals rather than on a predefined graph structure. The resulting graph can be processed by the graph convolutional layers to extract the more suitable features for emotion recognition. The features obtained are also tested in cross-subject scenarios. 

\subsection{\textbf{Transductive methods}}

Transductive methods \cite{vapnik200624} start from the hypothesis that the unlabeled test data are available in the training stage (no assumption about the distribution of the data, differently from the DA methods). The idea is that in several problems there is only a specific set of data (usually corresponding to the test set) to classify, and it is available at training time. Note that in standard ML approach the goal is to generalize on new unseen data, by contrast in transductive learning the goal is to correctly classify the test set only. The Transductive SVM (TSVM) is an example of a transductive method. Described in \cite{joachims1999transductive}, differently from classical SVMs that leverages only on labelled data, TSVMs exploit also unlabelled test data to find the best decision boundary between the classes. In other words, the target data is viewed as an additional set of information about the data. One of the main drawback of TSVM is that an estimation of the number of elements for each class in the test set is needed. Progressive TSVM (PTSVM, \cite{chen2003learning}) tries to resolve this problem progressively labeling the unlabeled data during the training instead of classifying it as a whole at the same time.

The only study collected in this review using explicitly transdutive methods is \cite{yang2020improving}, where the PTSVM generalisation power across several sessions of EEG data acquired from different subjects is validated.

\subsection{\textbf{Transfer Learning methods}}
Transfer Learning methods are based on the concepts of \textit{Domain} and \textit{Task}. Following the survey of Pan et al. \cite{pan2009survey}, a Domain can be defined as a set $D=\{F, P(X)\}$ where $F$ is a feature space and $P(X)$ is the marginal probability distribution of a specific dataset $X=\{x_1,x_2,\dots,x_n\} \in F$. 
Instead, a Task is a set $T=\{L, f\}$ where $L$ is a label space and $f$ is a predictive function $f$ usually learned by the data. For instance, $f(x_i)$ can be used to assign the predicted label to $x_i \in X$.
Therefore, $f$ can be equivalently viewed as the probability of a label $y$ given a data $x$, i.e. $p(y \in L \vert x \in X)$.

A Dataset of $n$ points can be defined as a set $S=\{(x_i\in X, y_i \in L)\}_{i=1}^n$.
Transfer learning wants to exploit the knowledge of a domain $D_A$ on a task $T_A$ to resolve the same or another task $T_B$ on another domain $D_B$.\\
By the definition of domain, it is straightforward that two domains $D_A=\{F_A, P(X_A)\}$ and $D_B=\{F_B, P(X_B)\}$ can be considered different if they differ in the feature spaces or in the marginal probability distributions. Obviously, the same holds for two Tasks $T_A=\{L_A, f_A\}$ and $T_B=\{L_B, f_B\}$.
More in details, the following cases can happen:
\begin{enumerate}
    \item $D_A=D_B$ and $T_A=T_B$: since the Tasks and the Domains are the same, this can be considered a classical Machine Learning Problem.
    \item $D_A \neq D_B$: $F_A \neq F_B$ or $F_A=F_B$ and $P(X_A) \neq P(X_B)$
    \item $T_A\neq T_B$: $L_A \neq L_B$ or $f_A \neq f_B$.
\end{enumerate}

The non-stationarity of the EEG signals between different subjects in an emotion classification problem can be viewed as a multi-domain problem where the data belonging to each subject are sampled from different Domains. Usually, given a pair of two different subjects $A$ and $B$, a common features space is assumed to be shared by the two domains (the EEG data representation), the conditional data distributions $P(L_A\vert X_A)=P(L_B\vert X_B)$ are assumed to be the same  while the marginal probability distributions are different on the available data, i.e. $P(X_A)=P(X_B)$. Therefore, minimising the non-stationarity of EEG signal should be viewed as reducing a discrepancy measure between several Domains. 

In the current literature, transfer learning strategies can be divided in three families:
\begin{itemize}
    \item \textbf{Unsupervised/semisupervised Domain Adaptation (DA)} methods,
    \item \textbf{Supervised DA Methods}, also known as \textbf{PreTraining, (PT)}  methods,
    \item \textbf{Domain Generalization (DG)} methods.
\end{itemize}

These families differ mainly in which data are processed during the learning stage. DA methods start from the hypothesis that data sampled from two different domains are available, called \textit{Source} Domain and \textit{Target} Domain respectively. The main difference between them is that, while a complete dataset $S_{Source}=\{(x_i, y_i)\}_{i=1}^n$ can be sampled from the Source domain, only feature data points $X_{Target}=\{x_j\}_{j=1}^m \in F_{Target}$ can be sampled from the Target one, without knowledge (unsupervised DA) or minimal knowledge (semi-supervised DA) of their real labels. Unsupervised DA methods can be viewed as transductive machine learning methods with the further hypothesis that the data come from two different distributions. Instead, PT methods work adapting a  model already trained on a known (Source) Domain to go toward a new (Target) domain. Since both features $X$ and labels $y$ of the new domain are known during the adapting stage, PT strategies are also known as \textit{supervised DA} methods. In contrast, DG methods rely on the hypothesis that $d \geq 2$ source domains together with their labeled samples are available, while any data from the Target domain is unknown.

DA and DG methods are getting a great deal of attention in the scientific literature in different contexts (e.g. image classification and voice recognition), and several proposals have been made until now. One trend of the literature is to adapt DA/DG methods originally proposed for a context to another one. For example, in \cite{zhou2020graph} methods to adapt DA strategies for image classification to EEG emotion classification are proposed. However, each context has its characteristics and peculiarities, making the transfer of a DA method from a task to another task not immediate. Several attempts were made by the scientific community to adapt well-established DA/DG methods in tasks involving the processing of EEG signals in the emotion recognition field.

\subsubsection*{\textbf{Domain Adaptation (DA) methods}}
Several DA methods relied on minimising the discrepance measures between the Source and the Target domain. In \cite{ganin2015unsupervised} these methods are categorised into shallow and deep DA.  As will be explained in the section (Proposed Taxonomy), an extension of this taxonomy is proposed in this paper, by adding the \textit{Source selection} methods:
\begin{itemize}
\item \textit{Source selection}: a subset of DA methods take into account that not all the training data can effectively be useful for the target space. Therefore, a selection of the training data is made, in order to avoid negative transfer. Several of these methods are also known as \textit{Instance weighting} \cite{jiang2008literature}, since they assign weights to the data; they can be made as a kind of preprocessing for the other methods;
\item \textit{Shallow DA}: the data representation is given a-priori. Only a mapping between the Source and Target representations is learned, without affecting the starting data representation;
\item \textit{Deep DA}: the data representation is learned as part of the DA strategy.
\end{itemize}

In the following part of this section, the studies according the above discussed ML approaches are reported.
 
\subsubsection*{\textit{Source selection}}
Source selection methods take into account that not all the training data can effectively be useful for the target space. In \cite{zhang2019individual} TrAdaBoost \cite{tradaboost} is used to score the source EEG data so that they does not negatively influence the training process. In other words, a small amount of labeled target data helps to vote on the usefulness of each of the available source data instance. As initial step, only the source subjects  closest to the target one are selected according to the MMD similarity and fed as auxiliary data to TrAdaBoost.
This family of methods can be also used as initial step of another DA method, for example in \cite{lin2017improving,zhou2020eeg} the similarity between source and target EEG data is measured using the Pearson Correlation Coefficient and the Average Frechet Distance respectively. In particular, in \cite{zhou2020eeg} only the closer EEG source data to the target data are fed to TCA together with the target one. Finally, the classification step is made by an Echo State Network (ESN, \cite{ozturk2007analysis}).  
In \cite{hua2021manifold}, Neighborhood Component Analysis (NCA, \cite{kenneth2020face}) is employed to learn the Mahalanobis distance between data and linearly project them into a subspace such that the classification accuracy is maximized and to reduce the dimensionality of the EEG features. The obtained features are then used with  Geodesic flow kernel for Unsupervised Domain Adaptation \cite{gong2012geodesic}.
In \cite{wang2021deep} (DMATN) data belonging to  the existing subjects are divided into several sub-source domains. Then, a set of sub-source are chosen as the most relevant with the target data. The proposed architecture combines together DAN and DANN to learn representation domain invariant representation.

\subsubsection*{Shallow DA methods}
Different strategies were proposed in literature, usually relied on one of following alternatives:
\begin{itemize}
    \item \textit{Target Space-Based (TSB)}: searching for a good transformation which directly maps Source data $S$ to the Target data $T$ space ($S \rightarrow T$);
    \item \textit{Shared Space-Based (SSB)}: searching for a good transformation which maps Source $S$ and Target $T$ data in a new shared space where the discrepancy between $S$ and $T$ is minimal ($S,T \rightarrow C$).
\end{itemize}
Once all the data are projected in a common space having the marginal distributions of the Domains close enough each others, common classification methods can be used to emotion recognition.

In Target space-based papers, \cite{fernando2013unsupervised} tried to align the source space toward the target one. Rather than using the data in their original feature spaces, the authors used PCA for a more robust and compact data representation. More specifically, two PCA projection matrices $Z_S$ and $Z_T$ are computed for the Source and the Target domain respectively. Therefore, a transformation matrix $M$ able to align the source space to the target one is searched by an optimisation problem, i.e. $$\arg\min\limits_{M}\vert \vert Z_S M - Z_T\vert \vert _F^2$$
This problem has a closed form solution in the form $M=Z_S^T Z_T$. Indeed, a similarity between the projected data can be computed.

In \cite{chai2017fast} ASFM is adopted for EEG-based Emotion Recognition.
However, as a large part of the domain adaptation strategies, \cite{chai2017fast} uses all the source subject as a whole as they belong to the same domain. In other words, the data belonging to different training subjects are viewed as an unique subject. 
Differently, in \cite{chai2018multi} (Multi-Subject Subspace Alignment, MSSA)  the ASFM strategy is applied to each source subject individually, then the projected data are fed to different for-subject classifiers.

Other data transformations have been investigated in the DA scenario for EEG emotion recognition, such as Robust Principal Component Analysis \cite{wright2009robust} in \cite{lin2019constructing}. RCA decomposes a set $X$ of data as $X=L+S$, where $L$ and $S$ are two superimposed matrices: the former is a low-rank matrix, the latter a sparse matrix. These matrices are computed resolving the following optimisation problem:
 $$\min_{L,S}\vert \vert L\vert \vert _* + \lambda \vert \vert S\vert \vert _1$$
 where $\vert \vert \cdot\vert \vert _*$ is the matrix nuclear norm, $\vert \vert \cdot\vert \vert _1$ the $l_1$ norm and $\lambda$ a weighting parameter. In \cite{lin2019constructing} a proposal to use RPCA to build a  Cross-Day emotion recognition model is made.
 
In \cite{li2019multisource} a method for  personalised handwriting recognition (Style Transfer Mapping, STM \cite{zhang2012writer}) is adapted for EEG emotion recognition task  to generalise across different subjects. In a nutshell, STM maps source data to target data by an affine transformation. The solution of the proposed problem is in closed form, so it can be easily computed. Few labeled target data are used to make a source data selection, so starting from the hypothesis that a small amount of labeled data are available. 
On the other side, ins Shared space-based the Maximum Mean Discrepancy (MMD,\cite{gretton2006kernel}) is one of the currectly most used discrepancy measure in DA/DG strategies.
In the original study, MMD is proposed to test if two probability distributions $p$ and $q$ are different or not. Formally, the authors show that in a Reduced Kernel Hilbert Space (RKHS) a discrepancy measure between the two distributions can be defined as
$$MMD(p,q)= \vert \vert \mathbb{E}_{X_S \sim p}(\phi(X_S)) - \mathbb{E}_{X_T \sim q}(\phi(X_T))\vert \vert ^2_H$$
where $\phi(\cdot)$ is an appropriate feature mapping. 
In \cite{gretton2006kernel} is proven that, in a RKHS, $MMD(p,q)$ is 0 if and only if the two distributions $p$ and $q$ are the same.\\
MMD can be empirical estimated as the difference between the averages of two data sampled from the two distributions projected in a RKHS. Therefore, considering $X_S$ and $X_T$ as two sets sampled from the Source and the Target domain respectively, empirical $MMD(X_S, X_T)$ can be expressed as:
$$MMD(X_S, X_T) = \vert \vert \frac{1}{\vert X_S\vert }\sum\limits_{i=1}^{\vert X_S\vert }\phi(\mathbf{x}^{(i)}_S) - \frac{1}{\vert X_T\vert }\sum\limits_{i=1}^{\vert X_T\vert }\phi(\mathbf{x}^{(i)}_T)\vert \vert _H^2$$ where $\mathbf{x}^{(i)}_S$ and $\mathbf{x}^{(i)}_T$ are elements of $X_S$ and $X_T$ respectively. In other words, having two samples from two different distributions, the distance between the two distributions can be estimate through the distance between two means of the samples projected in a RKHS.

Transfer Component Analysis (TCA, \cite{pan2010domain}) is one of the most used MMD-based DA method. In the original work, two different TCA versions were proposed: i) an \textit{unsupervised} version,  where a transformation of the data is found such that the data variance is maximally preserved reducing, at the same time, the MMD distance of the domains distributions, and ii) a \textit{supervised} one, where the data dependence with the training labels is taken into account.

An evaluation of the unsupervised TCA on EEG data for emotion recognition was made in \cite{zheng2015transfer}. Instead of using all the available EEG data, a random selection of a subset of samples from Source domain data was made during the evaluation strategy, letting out a subject as Target domain. In \cite{xue2020feature} TCA is tested on SEED dataset trying several desired dimensions for the feature space. Instead, in \cite{he2022cross} TCA is tested on self-made EEG data.

In \cite{long2013transfer} Transfer Sparse Coding (TSC) the MMD was exploited to find a sparse representation of image data sampled from different distribution. Sparse code representations are well-known data approximation obtained as linear combinations of elements in a set of basis functions. In a nutshell, a sparse coding method searches for a representative over-complete set of basis functions (a \textit{dictionary}) together with an encoding that best represent the data. In its simplest form, the sparse coding problem can be expressed as $$\min_{B,S}\vert \vert X- BS\vert \vert _F^2+ \lambda\sum\limits_{i=1}^n \vert \mathbf{s}_i\vert $$ 
where $X\in \mathbb{R}^{m \times n} $ is a matrix containing $n$ data points to approximate and $B\in \mathbb{R}^{m \times k}$ and $S\in \mathbb{R}^{k \times n}$ are the dictionary matrix and the encoding matrix respectively, where $k > m $ to ensure the over-completeness. The sparsity is induced by the second equation term on the coefficient matrix columns $\mathbf{s_i}$ and regulated through the hyperparameter $\lambda \in \mathbb{R}$. However, if $X$ is composed of data sampled from two different Domains (e.g., $X=[X_S \vert  X_T]$) the above formalisation does not take into account the differences between the marginal distributions. To deal with this problem,  \cite{long2013transfer} proposed to add  a further regularisation term to the objective function that takes into account the MMD distance between the different Domains of the input data.

In \cite{ni2021domain}, similarly,  a common dictionary between source and target domain is computed, but preserving the local information between samples and the discriminative knowledge between the domains exploiting the PCA and Fisher criteria \cite{fisher1992statistical}. This work required a little set of labelled data from the target domain, falling in the semi-supervised DA approaches.

While it is not specifically designed for Domain Adaptation, Kernel-PCA (KPCA,\cite{scholkopf1997kernel}) is often used in comparisons with several DA methods. In a nutshell, KPCA uses the kernel trick to project the data into a kernel space and then applying the PCA on the projected data. A comparison between Kernel-PCA and TCA for EEG emotion recognition is reported in \cite{zheng2015transfer}.
In \cite{chai2016unsupervised} (SAAE), a features transformation $\phi(\cdot)$ is computed through Kernel-PCA  maximising  the embedded data variance. Before the transformation, an autoencoder trained on data from both the Source and the Target Domains  was employed to preprocess the data. 
In \cite{zheng2016personalizing} TCA, KPCA, TSVM, and TPT are evaluated on the EEG-based emotional SEED dataset in a Leave-On-Subject-Out approach, while in \cite{lan2018domain} similar methods are tested on SEED and DEAP also for Cross-Dataset generalisation.

\subsubsection*{Deep DA methods}
In deep DA approaches, a feature data representation learning is embedded in the DA method. Instead of searching for a transformation of features given a priori, this is done by changing the feature space representation.

Deep DA methods can be further divided in:
\begin{itemize}
    \item \textit{Common Shared Space (CSS) methods}: Source and Target are projected in a new shared space
    \item \textit{Shared+Specific Spaces (SSS) methods}:  Source and Target are first projected in a unique shared space, then auxiliary more specific spaces are used.
\end{itemize}

As can be seen in Table \ref{best_performer}, CSS represents the category of studies exhibiting the best performance in terms of accuracy, at least in four of the cases considered. In \cite{tzeng2014deep} (Deep Domain Confusion, DDC) two identical networks are trained together, the former classifying data from the Source Domain, the latter adapting the distance between Source and Target domains using Target feature data. The combination of both the classification performance and the MMD is used as final loss.\\
\cite{zhang2019cross} uses DDC for cross-subject EEG emotion recognition.  The networks' architectures used are of type residual CNNs \cite{he2016deep}. To be fed to the CNNs, the EEG inputs are firstly transformed into Electrode-frequency Distribution Maps (EFDMs, \cite{wang2020emotion}). Results are validated with a Leave-One-Subject out CV approach. The authors of \cite{long2015learning} proposed a DA framework considering the general structure of a Convolutional Neural Network (CNN), that is usually composed by a sequence of convolutional layers followed by a fully-connected ones. 
The authors hypothesise that in a deep neural network the transition from general to the specific task features grows with the increasing of the network depth. Indeed, in a CNN,  while the initial convolutional layers learn general features, the final fully-connected ones learn domain specific features that are not transferable. Their proposed model (Deep Adaptation Network, DAN) deeply adapt the final  fully connected layers minimising the Multi-Kernel Maximum Mean Discrepancies (MK-MMD, \cite{zhu2017maximum}), a multiple kernel variant of MMD used as distribution discrepancy measurement. 
DAN was evaluated in EEG emotion recognition on SEED and SEED-IV in \cite{li2018cross}. In \cite{kuang2021cross} the proposed Multi-Spatial Domain Adaptation Network (MSDAN) aligns source and target domain considering the spatial relationships between the electrodes. This is done by using Graph Convolutional Layers and exploiting MMD distance in the resulting graph space. Differently from other works, \cite{kuang2021cross} uses data acquired in a Virtual Reality (VR) environment to generate stimuli, and the cross device problem is taken into account. One of the most used deep DA strategies is the Domain Adversarial Learning,  proposed in \cite{ganin2015unsupervised,ajakan2014domain,ganin2016domain}. The authors proposed an embedded problem formulation considering both the desired task and the discrepancy between the Source and the Target domain. The basic idea is to make the data distributions indistinguishable for an ad-hoc domain classifier. This can be made by a deep neural network model (Domain Adversarial Neural Network, DANN) that, for each input, predicts both the corresponding class and the belonging domain. In a nutshell,  DANN is composed of three main components: a feature extractor, a label predictor, and a domain classifier. Therefore, a learning process searches for a feature mapping maximising the class prediction performances and, at the same time, maximising the domain classification loss to make the feature distributions as similar as possible. 
DANN is evaluated in EEG emotion recognition task in \cite{jin2017eeg} on SEED. In \cite{li2018novel} BiDANN, a DANN variation is adopted for EEG emotion recognition, but considering the differences between the brain hemispheres, is proposed. In a nutshell, EEG data from the two hemispheres are processed separately: two different features mapping together with a domain discriminator are learned for the brain hemispheres, instead of only one feature mapping as in the original DANN formulation. Difference between the hemispheres in a DA approach is not dealt only by BiDANN; for instance, BiHDM \cite{li2018bi,li2020novel} uses two different RNN to code the data belonging to the two hemispheres, and also in this case a domain discriminator is used to mix up the features of the Source and the Target domain. In \cite{he2022adversarial} the authors propose a new DA method which is framed in the context of deep adversarial learning approaches. In particular a temporal convolutional network is used as encoder. Interestingly, the method is successfully evaluated in both cross-subject and cross-dataset. 
In \cite{ye2021cross,zhong2020eeg} domain adversarial approaches are used together with Graph Neural Networks (GNN, \cite{zhou2020graph}) as feature extractor.  In particular, \cite{ye2021cross} leverages on an attention mechanism \cite{niu2021review} focusing the learning stage on the alignment of the more changeling areas of the feature space. Performances are evaluated on SEED dataset. Instead, \cite{zhong2020eeg} proposes a node-wise domain adversarial training (NodeDAT) method to regularise the GNN output for better subject-independent performances. In EEG literature, Domain adversarial learning is widely used in several other studies for EEG data recognition, for example in \cite{tzeng2017adversarial,bao2020two,li2019regional,li2019domain,furukawa2021emotion, hwang2020subject}. In particular, in \cite{li2019regional} possible differences between several brain regions are also taken into account with a proposed attention module. 
In \cite{du2020efficient} (ATtention-based LSTM with Domain Discriminator, ATDD-LSTM) a domain discriminator in terms of LSTMs is presented to reduce the discrepance between the distributions. An attention-based encoder-decoder focuses on emotion-related helping the final classification probability estimation. An interesting adversarial approach was also investigated in \cite{wang2021prototype}. The proposed work exploits the Covariance Matrices between EEG data and Riemannian distances \cite{barbaresco2008innovative}. The work proposed a new kind of Neural Network (daSPDnet) able to  retain the intrinsic geometry information of the data. However, differently from the typical DA approach, a little set of labelled data belonging to the Target domain are required during the training process, resulting as semi-supervised DA method. A similar approach, also requiring a few of labelled target data, was proposed in \cite{li2021reducing}. Multi-source Domain Transfer Discriminative Dictionary Learning modeling (MDTDDL) is developed in \cite{gu2022multi}; the aim is to learn a joint subspace between source and target domains exploiting dictionary learning \cite{mairal2008supervised} methods.  DEEP and SEED are evaluated both in Cross-Subject and Cross-Session mode. In \cite{tzeng2017adversarial} Adversarial Discriminative Domain Adaptation (ADDA), a strategy to tackle the DA on an image classification task, was proposed. Differently from DANN, the ADDA basic idea consisted in building two different functions for the Source and the Target domains represented with two different encoders $E_S$ and $E_T$, respectively. $E_S$ is trained together with a classifier $C$ using labelled data from the Source domain. Then, through an adversarial learning procedure, $E_T$ is trained to map the Target domain data in the space of $E_S$ outputs. In this space, target data can be classified by $C$.
A similar idea was adapted in EEG emotion classification domain in \cite{luo2018wgan} (Wasserstein GAN Domain Adaptation, WGANDA), mixing together a pre-training stage and an adversarial training stage. More in detail, two generators for the source and target domain respectively are pre-trained to output two feature vectors of the same size. These vectors are considered as belonging to a shared feature space. An adversarial training step based on minimising the Wasserstein distance tunes the parameters of the generators such that the outputs match more closely as possible each other. The combined outputs are then used as input for an output classifier network. Inspired by the MMD optimization made in \cite{chai2016unsupervised}, in \cite{bao2020two}(TDANN) a two stage DA method is proposed. In the first stage, MMD is minimized training a CNN equipped with adaBN \cite{li2018adaptive}. To be fed to the CNN and to preserve spatial information, the EEG input signals are transformed into images \cite{bashivan2015learning,hwang2020learning}. In the second stage, a domain discriminator is used to further reduce the distance between the source and the target distributions. The method was evaluated in a leave-one-subject out cross validation framework. One of the main issue of the DANN networks is that no label is considered during the adversarial learning process, therefore the relationship between target data and the task-specific decision boundary during the distributions alignment is not taken into account. A DA method can confuse the distributions of the two domains by reducing the distance between them, resulting in a simple mixing of the samples of the two domains, leading the categories within each domain to not be distinguishable. Indeed, in DANN the decision boundary inside each domain is ignored.
In \cite{saito2018maximum} (Maximum Classifier Discrepancy, MCD), instead, the labels of the Source domain data are considered, helping to search a good task-specific decision boundaries between the classes. This is achieved by using different classifiers fed with the same inputs and evaluating the discrepancy. More in detail, two classifier $C_1$ and $C_2$ with the same characteristics are fed with input of feature generator $G$. $G$ can be fed with data $x$ coming from the source or the target domain. The output of $C_1$ and $C_2$ are the labels of the input $x$ returned by $G$. Before the training step, $C_1$ and $C_2$ start from different initial state, rising two different classifiers after the training. How much the two classifiers disagree on their predictions on the same input is defined \textit{discrepancy} by the authors. Indeed, the generator $G$ is trained to minimise the discrepancy (that is, project source and target data in the same space), while $C_1$ and $C_2$ are trained to maximise the discrepancy (so that the two classification boundaries are far from each other). The learned generator $G$ will be able to relocate the target domain data in the source space, but taking into account its most probable belonging class. Task-Specific Domain Adversarial Neural Network (T-DANN, \cite{ding2021eeg}) is an MCD similar model proposed for EEG emotion recognition. T-DANN adapts the conditional distribution between domains and, at the same time, adapts classification boundaries between classes exploiting MCD in conjunction with a domain discriminator. 
Instead, \cite{ning2021cross} deal with the excessive allignment problem exploiting a few-shot learning and attention mechanism approach. From a different point of view, \cite{wang2021cross} used Siamese Networks \cite{koch2015siamese} for evaluating the similarity between samples belonging to different domains. Siamese networks were originally proposed to to determine whether two inputs belong to the same category or not. In \cite{wang2021cross} the Siamese framework is converted to handle different domains. However, this method require a few of labelled data belonging to the Target domain. In \cite{tao2021multi} the authors propose a DA approach for EEG-based emotion recognition based on a multi-source co-adaptation framework (MACI). The proposed framework mainly takes advantage of correlation knowledge among several sources and features to build a proper objective function. The proposed method is compared with both standard (shallow) DA approaches and deep (CNN-based) approaches. Cross-subjects and cross-datasets evaluations are performed. Computational costs is a critical point of the proposed framework.\\
The authors of \cite{luo2021progressive} propose a novel approach which attempts to unify in an unique optimisation problem two standard DA approaches, instance reweighting (that we refer as source selection) and feature matching. This novel approach is named Progressive Low-Rank Subspace Alignment (PLRSA). In particular, instance reweighting is implemented by minimizing the Maximum Mean Discrepancy (MMD) distance and the TrAdaBoost algorithm, and feature matching by the Transfer Component Analysis (TCA). Importantly, a tiny amount of labeled target data is used to better exploit the source auxiliary data. The proposed method is evaluated in a both cross-subjects and cross-sessions scenario. The method is compared with five state-of-the-art DA methods. The results seem promising, however the time complexity is a little more expensive than related state-of-the-art methods.

Although several studies start from the hypothesis that a shared feature space is enough for DA, \textit{Shared+Specific Space (SSS)} methods go in different direction, believing that a single shared classifier built in a shared space still has poor performance for the never seen sessions/subjects. Notably, in these studies each subject/session available is considered as a single domain, and not as a whole. Hypothetically, EEG data representations can be splitted into shared emotional components, universal to all the subjects, and private components, specific to each subject.\\
Leveraging on this hypothesis, \cite{zhao2021plug} builds a shared encoder and private encoders for each source subject data to capture the subject-invariant emotional representations and private components, respectively. The obtained encoders are then used to build several emotion  classifiers. Finally, a new subject classifier using few data is built. All these classifier are built exploiting the shared encoder. A classifier fusion strategy is then applied to obtain the final classification result. However, the proposed framework requires few labelled target data, falling in the unsupervised DA category.
MEERNet \cite{chen2021meernet} considers different classifiers for each different domain (subject or session), preceded by a feature extractor shared by all the domains. Final classification is made averaging between domain-specific classifiers. 
Similarly, \cite{luo2021wasserstein} proposed a framework composed of a common feature extractor to map all the domains in a common subspace, a main task classifier or regressor, and private discriminators for each domain. The training is made reducing the Wasserstein distance between the marginal distribution of each source domain and target one in an adversarial way.
In \cite{chen2021ms} the authors propose a Multi Source-Marginal Distribution Adaptation (MS-MDA) algorithm for EEG emotion recognition. Also in this case, the key idea is that the final response is obtained by the average of the responses of  target-source specific classifiers, preceded by a common feature extractor. Notably, the authors explore the impact of different types of data normalisation on the performance of the proposed model. MS-MDA is compared with several standard DA methods and it has very promising results. Similarly, the authors of \cite{cao2021multi} propose Multi-Dource and Multi-Representation Adaptation (MSMRA), an approach with many similarities with respect MS-MDA  algorithm \cite{chen2021ms}. Both cross-subjects and cross-sessions evaluations are performed.

\subsubsection*{\textbf{Supervised DA (PreTraining) methods}}

In the supervised Domain Adaptation category, four studies were included in the reviews. In \cite{cimtay2020investigating} a pretrained version of  InceptionResnetV2 \cite{szegedy2017inception} is used as feature extractor for EEG data. The classification is made by a final network layer added to the InceptionResnetV2 module. 
Instead, \cite{pusarla2022learning} exploited DenseNet121 \cite{huang2017densely} as pre-trained model to build a new architecture fed with EEG data transformed in spectrogram images. \\
In \cite{wang2020emotion} a CNN model trained on different subjects and sessions taken from the SEED dataset is then re-trained on a small amount of data of a target subject taken from DEAP dataset to evaluate the cross-dataset emotion recognition performances.
In \cite{li2020foit} several classifiers trained on different data belonging to different subjects and sessions are ensembled together obtaining a final classifier suitable both for cross-sessions and cross-subjects EEG emotion recognition. 

\subsubsection*{\textbf{Domain Generalization (DG) methods}}

Finally, differently from classical domain adaptation methods, in Domain Generalization data from several domains are available, but no data from the test domain is observed during the training stage.
Differently from classical domain adaptation methods, data from several domains are available, but no data from the test domain is observed during the training stage \cite{muandet2013domain}.
Methods can be divided as:
\begin{itemize}
    \item \textit{shallow DG}: a data transformation is given a priori;
    \item \textit{deep DG}: the data rapresentation is learned as part of the DG strategy.
\end{itemize}

\textit{Shallow DG} methods share the same principles of shollow DA ones, building a shared space between domains letting the input data representation unchanged. Domain Invariant Component Analysis (DICA) \cite{muandet2013domain} searches for common features across several domains. Features data are transformed by a learned orthogonal transformation which minimizes the dissimilarity between a set of known domains preserving, at the same time, the relations between data features and their real labels. The authors also provided an unsupervised DICA version which did not take care of the class labels.

In \cite{ghifary2016scatter} Scatter Component Analysis (SCA) is proposed. The aim of the authors is to propose a method adapt both DG and DA requirements. SCA search for a data transformation where at the same time, i) the source and the target Domains are similar, ii) elements of the same class are similar, iii)  elements of different classes are well separated and, iv) the variance of the whole data is maximised. This is made introducing \textit{Scatter}, a measure closely related to MMD.
In \cite{ma2019reducing}, SCA and DICA are applied and evaluated on SEED dataset.

On the other side, in \textit{deep DG} methods include the data representation as part of the generalization strategy. In \cite{gonzalez2019eeg} data from similar subjects are used to train the same classifier. The similarity between data is computed through a clustering algorithm. This subset of similar subjects is used to train a final CNN classifier. In \cite{liu2021domain} a similar strategy is adopted, but for Domain Adaptation context.

\cite{hagad2021learning} join together BiDANN and Variational Autoencoder (VAE) obtaining a subject-invariant Bi-lateral Variational Domain Adversarial Neural Network (BiVDANN). VAEs are generative neural networks able to learn embedding of data constrained to a Gaussian distribution. As any classical autoencoder, a VAE is composed of an encoder network able to transforms data to an embedding space while a decoder network is able to reconstruct the original input from the embedding. In the proposed work, the learned features are further refined by domain adversarial training across different subjects to learn subject-independent features. Furthermore, to maximize dataset intercompatibility spectral topography data of the EEG signal are used as input. 

\begin{figure*}
\begin{center}
  \includegraphics[width=1 
  \textwidth]{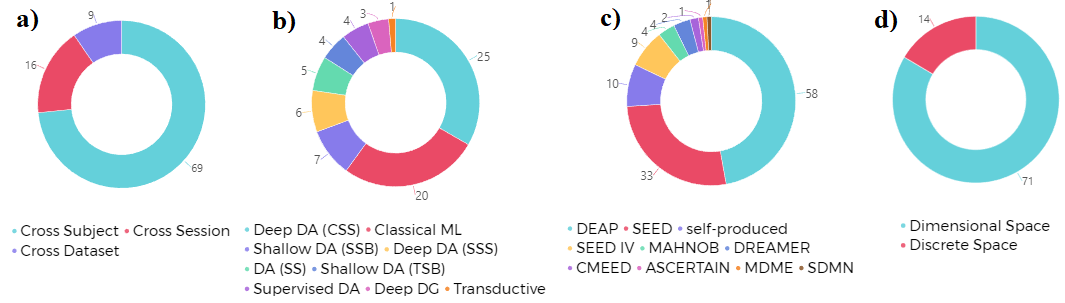}\\
  \caption{Pie charts for distribution of papers occurrences according to: a) generalization types, b) categories of the taxonomy, c) used datasets, d) emotional theories.}
  \label{statistics}
\end{center}
\end{figure*}

The pie charts in Figure \ref{statistics} show some statistics about the papers included in the survey. First of all, it is evident that almost three quarters of the studies surveyed (73.4 \%) focus on a cross-subject mode of generalisation, while cross-session studies account for only 17 \% and only 9 \% operate a cross-dataset mode of generalisation. Graph b) shows the percentage distribution according to the proposed taxonomy. Looking at this graph, it is evident that the majority of generalisation studies are moving towards the use of Deep Domain Adaptation (CSS) (33.33 \%), at the expense of more traditional approaches, which still retain 26.67 \%. This is followed by Shallow DA (SSB) approaches (9.33 \%), Deap DA (SSS) (8 \%), Source Selection DA (6.67 \%), Shallow DA (TSB) and Supervised DA (5.33 \%), Deep DG (4 \%) and finally Transductive (1.33 \%). \\
The pie chart in Figure \ref{statistics}.c) shows the number of times each EEG dataset is exploited in the reviewed literature. The mainly used datasets are SEED (45.8 \%) and DEAP (27.5 \%). This is followed by 8.3 \%   of studies that propose their own self-produced dataset, while of the other datasets available in the literature only SEED IV stands out (at 7.5 \%), which is interesting in that it adopts a discrete space of four emotions for classification (happy, sad, fear and neutral). Each of the other datasets do not exceed 5 \% (MAHNOB \cite{soleymani2011multimodal} and DREAMER \cite{katsigiannis2017dreamer} (3.33 \%), CMEED \cite{zhao2018frontal} (1.7 \%), ASCERTAIN \cite{subramanian2016ascertain}, MDME \cite{shenoy2006towards} and SDMN \cite{lin2010eeg} (0.8 \%)).

Finally, Figure \ref{statistics}.d) offers an interesting statistic about the interest of the authors of the studies examined in the various perspectives of emotion representation. As already mentioned in section 1.3, the two dominant perspectives, and the only ones considered in the literature examined, are those based on categorical and dimensional models. More than 80 per cent of the works are based on a representation of emotions, and then their subsequent classification in terms of valence and arousal (and only in one case also dominance \cite{wang2021prototype}). 



\begin{table*}[h!]
\caption{Reviewed studies on generalization strategies for emotion recognition. Datasets used, classifiers, evaluation strategy and type of generalization (i.e. intersubjects, cross sessions and cross datasets) are presented for each entry in the table. (sp = self produced, nl = not labelled; for the other abbreviations see section \ref{acronyms}).}
\label{complete_table}
\centering
\footnotesize 
\renewcommand{\arraystretch}{0.8}
\setlength{\tabcolsep}{4.5pt}
\scalebox{0.6}{

\begin{tabular}{c||c|c|c|c|c|c|c|}
   \multirow{2}{*}{Classifier Category} & \multirow{2}{*}{Study} & 
   \multirow{2}{*}{Dataset} & \multirow{2}{*}{Classifier} & \multirow{2}{*}{Evaluation Strategy} &
   \multirow{2}{*}{Cross Subject} &
   \multirow{2}{*}{Cross Session} &
   \multirow{2}{*}{Cross Dataset}\\
   \multirow{2}{*}{} & & & & & & &   \\
   
\toprule
\toprule

& \cite{yin2017cross} & DEAP & 
TRFE & LOO & X & & \\
\cmidrule[0.5pt](r{1pt}l{0pt}){2-8}
& \cite{rayatdoost2018cross} & DEAP, MAHNOB, sp & 
RF & LOO & X & & X \\
\cmidrule[0.5pt](r{1pt}l{0pt}){2-8}
& \cite{li2018exploring} & DEAP, SEED & SVM & LOO & X & & \\
\cmidrule[0.5pt](r{1pt}l{0pt}){2-8}
& \cite{song2018eeg} & SEED, DREAMER & DGCNN & LOO & X & &\\
\cmidrule[0.5pt](r{1pt}l{0pt}){2-8}
& \cite{yang2019multi} & DEAP, SEED & SVM & LOO & X & & \\
\cmidrule[0.5pt](r{1pt}l{0pt}){2-8}
& \cite{jiang2019cross} & DEAP, sp & SBS & LOO & X & &\\
\cmidrule[0.5pt](r{1pt}l{0pt}){2-8}
& \cite{cai2019multiple} & DEAP & TRFE & LOO & X & & \\
\cmidrule[0.5pt](r{1pt}l{0pt}){2-8}
CLASSICAL ML & \cite{li2019variational} & DEAP, SEED & VAE-LSTM & LOO & X & & \\
\cmidrule[0.5pt](r{1pt}l{0pt}){2-8}
& \cite{pandey2019subject} & DEAP & SVM & LOO & X & & \\
\cmidrule[0.5pt](r{1pt}l{0pt}){2-8}
& \cite{arevalillo2019combining} & DEAP, MAHNOB, DREAMER & SVM & LOO & X & & \\
\cmidrule[0.5pt](r{1pt}l{0pt}){2-8}
& \cite{zhang2020eeg} & DEAP, MAHNOB & RFE & LOO & X & & \\
\cmidrule[0.5pt](r{1pt}l{0pt}){2-8}
& \cite{liu2020subject} & SEED & DECNN & LOO & X & & \\
\cmidrule[0.5pt](r{1pt}l{0pt}){2-8}
& \cite{li2020latent} & DEAP, SEED & VAE-LSTM & LOO & X & &\\
\cmidrule[0.5pt](r{1pt}l{0pt}){2-8}
& \cite{chen2021personal} & SEED & SVM & LOO & X & & \\
\cmidrule[0.5pt](r{1pt}l{0pt}){2-8}
& \cite{fernandez2021cross} & SEED & SVM & LOO & X & & \\
\cmidrule[0.5pt](r{1pt}l{0pt}){2-8}
& \cite{tian2021personality} & ASCERTAIN & BiLSTM & LOO & X & & \\
\cmidrule[0.5pt](r{1pt}l{0pt}){2-8}
& \cite{li2021cross} & SEED, SEED IV & SOGNN & LOO & X & & \\
\toprule
TRANSDUCTIVE & \cite{yang2020improving} & SEED, sp & PTSVM & LOO & & X & \\
\toprule
& \cite{lin2017improving} & sp & GNB & ASI & & X & \\
\cmidrule[0.5pt](r{1pt}l{0pt}){2-8}
& \cite{zhang2019individual} & DEAP & SVM & LOO & X & & \\
\cmidrule[0.5pt](r{1pt}l{0pt}){2-8}
DOMAIN ADAPTATION & \cite{zhou2020eeg} & DEAP & ESN & LOO & X & & \\
\cmidrule[0.5pt](r{1pt}l{0pt}){2-8}
(SOURCE SELECTION) & \cite{hua2021manifold} & DEAP & NCA & LOO & X & & \\
\cmidrule[0.5pt](r{1pt}l{0pt}){2-8}
& \cite{wang2021deep} & SEED & DMATN & LOO & X & & \\
\toprule
& \cite{chai2017fast} & SEED & ASFM & LOO & X & X & \\
\cmidrule[0.5pt](r{1pt}l{0pt}){2-8}
SHALLOW DA (TSB) & \cite{chai2018multi} & SEED & MSSA & LOO & X & & \\
\cmidrule[0.5pt](r{1pt}l{0pt}){2-8}
& \cite{lin2019constructing} & MDME, SDMN & RPCA & ASI & & X & \\
\cmidrule[0.5pt](r{1pt}l{0pt}){2-8}
& \cite{li2019multisource} & SEED & STM & LSO & X & &\\
\toprule
& \cite{zheng2015transfer} & SEED & TCA & LOO & X & & \\
\cmidrule[0.5pt](r{1pt}l{0pt}){2-8}
& \cite{chai2016unsupervised} & SEED & SAAE & SU2SU, SE2SE, LOO & X & X & \\
\cmidrule[0.5pt](r{1pt}l{0pt}){2-8}
& \cite{zheng2016personalizing} & SEED & TPT & LOO & X & & \\
\cmidrule[0.5pt](r{1pt}l{0pt}){2-8}
SHALLOW DA (SSB) & \cite{lan2018domain} & DEAP, SEED & MIDA & LOO & X & & X \\
\cmidrule[0.5pt](r{1pt}l{0pt}){2-8}
& \cite{xue2020feature} & SEED & TCA & LOO & X & & \\
\cmidrule[0.5pt](r{1pt}l{0pt}){2-8}
& \cite{ni2021domain} & DEAP, SEED & DASRC & LOO & & X & X \\
\cmidrule[0.5pt](r{1pt}l{0pt}){2-8}
& \cite{he2022cross} & sp & TCA & HO & & X & \\
\toprule
& \cite{jin2017eeg} & SEED & DANN & LOO & X & & \\
\cmidrule[0.5pt](r{1pt}l{0pt}){2-8}
& \cite{li2018cross} & SEED, SEED IV & DAN & LOO & X & & \\
\cmidrule[0.5pt](r{1pt}l{0pt}){2-8}
& \cite{li2018novel} & SEED & BiDANN & LOO & X & & \\
\cmidrule[0.5pt](r{1pt}l{0pt}){2-8}
& \cite{luo2018wgan} & DEAP & WGANDA & LOO & X & & \\
\cmidrule[0.5pt](r{1pt}l{0pt}){2-8}
& \cite{zhang2019cross} & SEED & DDC & LOO & X & & \\
\cmidrule[0.5pt](r{1pt}l{0pt}){2-8}
& \cite{li2019regional} & SEED & R2G-STNN & LOO & X & &\\
\cmidrule[0.5pt](r{1pt}l{0pt}){2-8}
& \cite{li2019domain} & DEAP, SEED & nl & LOO, SU2SU, O2OSE & X & X & \\
\cmidrule[0.5pt](r{1pt}l{0pt}){2-8}
& \cite{li2020novel} & SEED, SEED IV, MPED & BiHDM & LOO & X & & \\
\cmidrule[0.5pt](r{1pt}l{0pt}){2-8}
& \cite{zhong2020eeg} & SEED, SEED IV & RGNN & LOO & X & & \\
\cmidrule[0.5pt](r{1pt}l{0pt}){2-8}
& \cite{bao2020two} & SEED, sp & TDANN & LOO & X & X & \\
\cmidrule[0.5pt](r{1pt}l{0pt}){2-8}
DEEP DA (CSS) & \cite{hwang2020subject} & SEED, CMEED & A-DNN & LOO & X & & \\
\cmidrule[0.5pt](r{1pt}l{0pt}){2-8}
& \cite{du2020efficient} & DEAP, SEED, CMEED & ATDD-LSTM & LOO & X & X & \\
\cmidrule[0.5pt](r{1pt}l{0pt}){2-8}
& \cite{kuang2021cross} & sp & MSDAN & LOO & X & & \\
\cmidrule[0.5pt](r{1pt}l{0pt}){2-8}
& \cite{furukawa2021emotion} & SEED & nl & LOO & X & & \\
\cmidrule[0.5pt](r{1pt}l{0pt}){2-8}
& \cite{li2021reducing} & SEED & nl & LOO & X & & \\
\cmidrule[0.5pt](r{1pt}l{0pt}){2-8}
& \cite{ding2021eeg} & SEED & TDANN & SU2SU & X & & \\
\cmidrule[0.5pt](r{1pt}l{0pt}){2-8}
& \cite{ning2021cross} & DEAP, SEED & SDA-FSL & LOSO & X & & X \\
\cmidrule[0.5pt](r{1pt}l{0pt}){2-8}
& \cite{tao2021multi} & DEAP, SEED & MACI & LOO & X & & X \\
\cmidrule[0.5pt](r{1pt}l{0pt}){2-8}
& \cite{luo2021progressive} & DEAP, SEED & PLRSA & LOO, SU2SU & X & X &\\
\cmidrule[0.5pt](r{1pt}l{0pt}){2-8}
& \cite{he2022adversarial} & DEAP, DREAMER & AD-TCN
& LOO & X & & X \\
\cmidrule[0.5pt](r{1pt}l{0pt}){2-8}
& \cite{li2018bi} & SEED & BiHDM & LOO & X & &\\
\cmidrule[0.5pt](r{1pt}l{0pt}){2-8}
& \cite{gu2022multi} & DEAP, SEED & MDTDDL & LOO & X & & X \\
\toprule
\toprule
\end{tabular}
}

\end{table*}

\begin{table*}[h!]
\ContinuedFloat 
\label{summary}
\centering
\footnotesize 
\renewcommand{\arraystretch}{0.8}
\setlength{\tabcolsep}{4.5pt}
\scalebox{0.6}{


\begin{tabular}{c||c|c|c|c|c|c|c|}
   \multirow{2}{*}{Classifier Category} & \multirow{2}{*}{Study} & 
   \multirow{2}{*}{Dataset} & \multirow{2}{*}{Classifier} & \multirow{2}{*}{Evaluation Strategy} &
   \multirow{2}{*}{Cross Subject} &
   \multirow{2}{*}{Cross Session} &
   \multirow{2}{*}{Cross Dataset}\\
   \multirow{2}{*}{} & & & & & & &   \\
   
\toprule
\toprule
& \cite{zhao2021plug} & SEED & PPDA & LOO & X & & \\
\cmidrule[0.5pt](r{1pt}l{0pt}){2-8}
& \cite{chen2021meernet} & SEED, SEED IV & MEERNet & LOO & X & X & \\
\cmidrule[0.5pt](r{1pt}l{0pt}){2-8}
DEEP DA (SSS) & \cite{liu2021domain} & DEAP, sp & DASC & LOO & X & & \\
\cmidrule[0.5pt](r{1pt}l{0pt}){2-8}
& \cite{chen2021ms} & SEED & MS-MDA & LOO & X & X & \\
\cmidrule[0.5pt](r{1pt}l{0pt}){2-8}
&
\cite{luo2021wasserstein} & SEED & wMADA & LOO & X & &\\
\toprule
& \cite{cimtay2020investigating} & DEAP, SEED, sp & nl & LOO & X & & X\\
\cmidrule[0.5pt](r{1pt}l{0pt}){2-8}
SUPERVISED DA & \cite{wang2020emotion} & DEAP, SEED & RCNN & LOO & X & & X\\
\cmidrule[0.5pt](r{1pt}l{0pt}){2-8}
& \cite{pusarla2022learning} & DEAP, SEED & Densenet & LOO & X & &\\
\toprule
DEEP DG &
\cite{ma2019reducing} & SEED & DG-DANN, DResNet & LOO & X & &\\
\cmidrule[0.5pt](r{1pt}l{0pt}){2-8}
& \cite{hagad2021learning} & DEAP, SEED & BiVDANN & LOO & X & &\\
\toprule
\toprule
\end{tabular}

}

\end{table*}

\begin{table*}[h!]
\caption{The most representative studies according to their classification accuracy, categorised by EEG dataset (SEED, DEAP, others) and by number and type of classes considered. DIM = Dimensional; DIS = Discrete; sp = self-produced.}
\label{best_performer}
\centering
\footnotesize 
\renewcommand{\arraystretch}{1.5}
\setlength{\tabcolsep}{1pt}
\scalebox{0.7}{

\begin{tabular}{c||c|c|c|c|c|c|c|}
   \multirow{2}{*}{Proposed} & \multirow{2}{*}{Study} &
   \multirow{2}{*}{Dataset} &
   \multirow{2}{*}{Reference} &
   \multirow{2}{*}{\#Classes} &
   \multirow{2}{*}{Accuracy} \\
   \multirow{2}{*}{} category & & & Theory & &  \\

\toprule
\toprule

SUPERVISED DA & \cite{cimtay2020investigating} 
& DEAP & DIM (VAL) &  \#2 (LV/HV) & 72.81 ± 5.07 \\
\cmidrule[0.5pt](r{1pt}l{0pt}){3-6}
& & Other & DIM (VAL) & \#2 (LV/HV) & 81.80 ± 10.92 \\
\toprule
DA - SOURCE SELECTION &
\cite{zhou2020eeg} & DEAP & DIM (VAL) &  \#3 (LV/MV/HV) & 68.06 ± 10.93 \\
\toprule
DEEP DA (SSS) &
\cite{liu2021domain} & DEAP & DIM (VAL-ARO) &  \#2 (LV/HV) & 73.90 ± 13.50 (VAL) \\
& & & & \#2 (LA/HA) & 68.80 ± 11.20 (ARO)\\
\toprule
DEEP DA (CSS) &
\cite{li2019domain} & DEAP & DIM (VAL-ARO) &  \#4 (LALV-HALV- & 62.66 ± 10.45  \\
& & & &   LAHV-HAHV) & \\
\toprule
DEEP DA (CSS) &
\cite{ning2021cross} & SEED & DIM (VAL) &  \#2 (LV/HV) & 97.66 ± 14.46 \\
\toprule
DEEP DA (CSS) &
\cite{du2020efficient} & SEED & DIM (VAL) &  \#3 (LV/MV/HV) & 90.92 ± 1.05  \\
\cmidrule[0.5pt](r{1pt}l{0pt}){3-6}
& & Other & DIM (VAL-ARO) &  \#2 (LV/HV) & 94.21 ± 5.88 (VAL) \\
& & & & \#2 (LA/HA) & 88.03 ± 6.32  (ARO)\\
\toprule
DEEP DA (CSS) &
\cite{bao2020two} & Other & DIS (HAPPY, SAD &  \#2 (JOY/SADNESS) & 83.79 ± 1.55 (JOY/SAD)  \\
& & & FEAR, ANGER) &   \#2 (JOY/ANGER) & 84.13 ± 1.37 (JOY/ANGER)\\
& & & & \#2 (JOY/FEAR) & 81.72 ± 1.30 (JOY/FEAR)\\
\toprule
CLASSICAL ML &
\cite{li2021cross} & Other & DIS (HAPPY, SAD &  \#4 (HAPPY/SAD/ & 75.27 ± 8.19  \\
& & & FEAR, NEUTRAL) &   FEAR/NEUTRAL) & \\
\toprule

\toprule
\toprule
\end{tabular}
}

\end{table*}


Studies identified as best performers in terms of mean classification accuracy are proposed in Table \ref{best_performer}. Only cross-subject studies were considered in this selection, being the statistically most significant percentage among all the generalisation studies surveyed. The ten best results were identified considering as many classification issues. Each issue is defined by considering the number and type of classes identified (binary and ternary on valence and arousal, quaternary on the two-dimensional valence-arousal plane, binary and quaternary on discrete dimensions) and the dataset adopted (considering three possibilities: DEAP, SEED, other). Only studies reporting both mean accuracy and standard deviation were included in the performance assessment.

\section{Discussion}
\label{discussion}

To date, no robust electroencephalographic patterns are recognized in scientific literature for correlating with emotional states. 
Some studies base their results on the asymmetry of scalp activations, but several theories based on statistical samples that are not yet particularly large still coexist \cite{demaree2005brain, coan2003state, davidson1984hemispheric}.
When aiming for a generalization goal 
in EEG-based Emotion Recognition, Transfer Learning methods are becoming more and more established in the literature.
Domain Adaptation methods (Deep DA (CSB), Shallow DA (SSB), Source Selection DA, Deep DA (SSS), Shallow DA (TSB) and Supervised DA) exceed 60 \% of the total surveyed studies and exhibit very high accuracy performances in the table of best performers (see Table \ref{best_performer}).
In particular, Deep DA (Common Shared Space) is used by five best performers studies. This could be also due to the current massive use of Deep DA (Common Shared Space) in the literature. Indeed, one third of the surveyed studies (Figure \ref{statistics}.d) belongs to this category.

However, a still substantial percentage of works (27.4 \%) belongs to the Classical ML category.
An emblematic case in this context is \cite{li2021cross}, namely the best performer in the classification issue on SEED IV with four discrete classes. This is an interesting study based on a self-organized graph construction module. This solution can be considered as a peculiar implementation of the well established adaptive filters strategy, when the generalization goal is pursued by customising the network to the current input. Conversely, the DA strategies make the data from different domains more homogeneous by means of appropriate transformations.
The different impact between DA and adaptive filters approaches can be better appreciated by making a comparison between the previous study and \cite{zhong2020eeg}. 
Both studies address the problem of four-class classification on the same dataset by using a pipeline based on graphs and deep networks. In the first case, an adaptive graph is used without any DA methods, while the second study makes use of a (nonadaptive) 
graph approach in combination with Domain Adaptation techniques. Even though they use different approaches, the reported accuracy performances are comparable.
This suggests how the dynamic search for feature extraction procedures represents an interesting frontier for future studies in this area, not excluding the potential of using this approach in combination with DA/DG techniques.

Another point to take into account is that the proliferation of EEG acquisition devices on the market is not always coupled with consistency in terms of quality between the various devices (considering electrode type and positioning, interference shielding and signal-to-noise ratio, amplification strategies, etc). 
A comparison among different studies must take into account the quality of EEG instrumentation used. 
The IEC 60601-2-26 standard applies to basic safety and essential performance of electroencephalographs used in a clinical environment.  Among the requirements, the minimum overall signal quality for an electroencephalographic device to be considered acceptable is defined \cite{goldsack2020verification}. Even if IEC 60601-2-26 is a standard specifically developed for clinical purposes, it is nowadays the only available standard for EEG instrumentation quality certification. In the future, it is desirable for research to be increasingly based on certified instruments.
However, an encouraging trend emerges from the most recent public datasets. They are all based on standardized equipment: (i) Neuroelectrics Enobio 8 in the case of LUMED \cite{cimtay2020investigating}, (ii)  NuAmp Neuroscan in the case of CMEED \cite{du2020efficient}, and (iii) gtec.HIamp in the case of the dataset produced by \cite{bao2020two}).

A further concern in the use of public datasets is its underlying theoretical background, often acritically accepted by the scientists. Many studies validate the same machine learning algorithm on different datasets although the targeted psychic phenomena are radically different. Indeed, each dataset leverages on a specific theory of emotions and related experimental setup of emotion elicitation. For instance, DEAP is based on a dimensional approach and SEED IV on discrete one. 
The discrete theory is based on the assumption that there are basic emotions that have evolved through natural selection \cite{tenhouten2017primary}. In this vein, close to the Darwinian tradition, Ekman's theory identifies six basic emotions that would be universal and innate: anger, disgust, fear, happiness, sadness and surprise \cite{ekman1999basic}. After him, Plutchik, identifies eight basic emotions (anger, anticipation, joy, trust, fear, surprise, sadness and disgust) and arranges them on a wheel model \cite{suhaimi2020eeg}. 
In contrast, the dimensional theory expresses emotions in a continuous two-dimensional (valence-arousal) or three-dimensional (valence-arousal-dominance) space. While valence measures levels of pleasantness (happy vs. sad) of an emotion, arousal identifies degrees of excitement or motivational activation. In the three-dimensional model, the dominance dimension is added to valence and arousal, where the dominance evaluates emotions on a scale between submission and empowerment \cite{torres2020eeg}.
The underlying assumption of discrete approach is that few fundamental emotions are mediated by associated dedicated neural circuits, with a large innate (hardwired) component.
Only two main brain networks are recognized by the dimensional approach \cite{posner2009neurophysiological}.
The two theoretical approaches identify two different phenomena also at a neurophysiological level, with peculiar spatial signal features.
Finally, at present, the available public datasets do not adopt an established practice of psychological screening of the subjects involved. In general, studies on EEG-based emotion assessment could benefit from administering psychometric questionnaires to participants. Indeed, psychological data could help to understand individual differences in emotional response, leading to clustering of subjects \cite{liu2021domain}.
Recently, unsupervised clustering based on large datasets is emerging as a promising strategy for empirical identification of personality types \cite{gerlach2018robust}. Meanwhile, correlations have been found between personality types and EEG patterns \cite{li2020eeg}.
Moreover, prior psychological assessments allow to manage bias due to individual traits or states.
The introduction of psycho-metric tests and assessments during the production of upcoming datasets could lead to a much more fruitful use of data in support of generalization.

\section{Conclusion}
\label{conclusion}

A systematic literature review collecting papers on machine learning strategies to pursue (cross-subjects and cross-sessions) generalizability in EEG-based emotion recognition was carried out. Among the 418 articles retrieved from Scopus, IEEE (Institute of Electrical and Electronics Engineers) Xplore, and PubMed databases, 75 papers resulted eligible. A taxonomy of the studies employing ML method was proposed.

The studies with the best results in terms of average classification accuracy were identified, and the ten best results considering as many classification problems were highlighted. 
An interesting perspective based on self-organized graph construction modules emerged as peculiar strategy. This suggests how the adaptive feature extraction procedures represent an interesting frontier for future studies in this area, not excluding the potential of using this approach in combination with DA/DG techniques.

Future research on EEG-based emotion assessment could also benefit from administering psychometric questionnaires to participants in order to conduct a psychological screening of the experimental sample. This could help to understand individual differences in emotional responses, leading to clustering of subjects also taking into account the different subjects' personality.

\section*{Acronyms}
\label{acronyms}

\textbf{A-DNN}  - Adversarial Deep Neural Network\\
\textbf{AD-TCN} - Adversarial Discriminative Temporal Convolutional Network\\
\textbf{ASFM} - Adaptive Subspace Feature Matching\\
\textbf{ASI} -  Add-Session-In\\
\textbf{ATDD-LSTM} - Attention-based LSTM\\
\textbf{BiDANN} - Bi-hemispheres DANN\\
\textbf{BiHDM} - Bi-Hemispheric Discrepancy Model\\
\textbf{BiLSTM} - Bidirectional LSTM\\
\textbf{BiVDANN} - Bi-lateral Variational Domain Adversarial Neural Network\\
\textbf{CSS} - Common Shared Space\\
\textbf{DAN} - Deep Adaptation Network\\
\textbf{DANN} - Domain Adversarial Neural Network\\
\textbf{DASC} - Domain Adaptation Subject Clustering\\
\textbf{DASRC} - Domain Adaptation Sparse Representation Classifier\\
\textbf{DDC} - Deep Domain Confusion\\
\textbf{DECNN} - Dynamic Empirical Convolutional Neural network\\
\textbf{DGCNN} - Dynamical Graph Convolutional Neural Networks\\
\textbf{DG-DANN} - Domain Generalization DANN\\
\textbf{DResNet} - Domain Residual Network\\
\textbf{ESN} - Echo State Network\\
\textbf{GNB} - Gaussian Naïve Bayes\\
\textbf{HO} - Hold Out\\
\textbf{LOO} - Leave One Out\\
\textbf{LSTM} - Long short-term memory\\
\textbf{MACI} - Multi-Source Co-adaptation Correlation Information\\
\textbf{MDTDDL} - Multi-source Domain Transfer Discriminative Dictionary Learning modelling\\
\textbf{MEERNet} - Multi-Source EEG-based Emotion Recognition Network\\
\textbf{MIDA} - Maximum Independence Domain Adaptation\\
\textbf{MSDAN} - Multi-Spatial Domain Adaptation Network\\
\textbf{MS-MDA} - Multi Source-Marginal Distribution Adaptation\\
\textbf{MSSA} - Multi-Subject Subspace Alignment\\
\textbf{Na} - not available\\
\textbf{NCA} = Neighborhood Component Analysis\\
\textbf{O2OSE} -  ONE-TO-ONE-SESSION\\
\textbf{PLRSA} - Progressive Low-Rank Subspace Alignment\\
\textbf{PPDA} - Plug-and-Play Domain Adaptation\\
\textbf{R2G-STNN} - Regional To Global Spatial-Temporal Neural Network\\
\textbf{RCNN} - Residual CNN\\
\textbf{RF} - Random Forest\\
\textbf{RFE} - Recursive Feature Elimination\\
\textbf{RGNN} - Regularized Graph Neural Network\\
\textbf{RPCA} - Robust Principal Component Analysis\\
\textbf{SAAE} - Subspace Alignment Auto Encoder\\
\textbf{SBS} - Sequential Backward Selection\\
\textbf{SDA-FSL} - Single-Source Domain Adaptive Few-Shot Learning Network\\
\textbf{SE2SE} - session-to-session\\
\textbf{SOGNN} - Self-Organized Graph Neural Network\\
\textbf{sp} - self-produced \\
\textbf{SSB} - Shared Space-Based\\
\textbf{SSS} - Shared+Specific Spaces\\
\textbf{STM} - Style Transfer Mapping\\
\textbf{SU2SU} - subject-to-subject\\
\textbf{SVM} - Support Vector Machine\\
\textbf{TCA} - Transfer Component Analysis\\
\textbf{TDANN} - Two-Level Domain Adaptation Neural Network\\
\textbf{TPT} - Transductive Parameter Transfer\\
\textbf{TRFE} - Transferable Recursive Feature Elimination\\
\textbf{TSB} - Target Space-Based \\
\textbf{VAE} - Variational Auto Encoder\\
\textbf{WGANDA} - Wasserstein Generative Adversarial Network Domain Adaptation\\
\textbf{wMADA} - Wasserstein-Distance-based Multi-Source Adversarial Domain Adaptation\\

\bibliography{__bibliografia}
\end{document}